\pdfoutput=1

\documentclass[11pt]{article}

\usepackage{ACL2023}

\usepackage{times}
\usepackage{latexsym}

\usepackage[T1]{fontenc}

\usepackage[utf8]{inputenc}

\usepackage{microtype}

\usepackage{inconsolata}

\usepackage{hyperref}
\usepackage{url}
\usepackage{graphicx}
\usepackage{xspace}
\usepackage[most]{tcolorbox}
\usepackage{xcolor}
\usepackage[detect-weight=true, detect-inline-weight=math]{siunitx}

\usepackage{booktabs}
\usepackage{multirow}
\usepackage{subcaption}
\usepackage{colortbl}
\usepackage{enumitem}

\definecolor{lightgrey}{HTML}{E7E7E7}

\definecolor{mypink}{HTML}{f8adb1}
\definecolor{mygreen}{HTML}{c7f0b1}
\definecolor{myyellow}{HTML}{ffecaf}
\definecolor{myblue}{HTML}{e0fbff}

\newcommand\eg[0]{\textit{e.g.}}
\newcommand\ie[0]{\textit{i.e.}}

\newcommand{\OurMethod}{{\textsc{Middleware}}\xspace}

\newcommand{\Freebase}{{\textsc{Freebase}}\xspace}

\newcommand{\WebQSP}{{\textsc{WebQSP}}\xspace}
\newcommand{\ComplexQ}{{\textsc{ComplexWebQ}}\xspace}
\newcommand{\GraphQ}{{\textsc{GraphQ}}\xspace}
\newcommand{\MetaQ}{{\textsc{MetaQA}}\xspace}
\newcommand{\Spider}{{\textsc{Spider}}\xspace}
\newcommand{\Bird}{{\textsc{Bird}}\xspace}

\newcommand{\KBQAAgent}{{\textsc{KBQA-Agent}}\xspace}
\newcommand{\WikiSQL}{{\textsc{WikiSQL}}\xspace}
\newcommand{\GrailQ}{{\textsc{GrailQA}}\xspace}

\newenvironment{entityfont}{\fontfamily{qcr}\selectfont\footnotesize}{\par}
\newenvironment{relationfont}{\fontfamily{qcr}\selectfont\footnotesize}{\par}
\DeclareTextFontCommand{\textentity}{\entityfont}
\DeclareTextFontCommand{\textrelation}{\relationfont}

\newenvironment{remark}[1][Remark]{\begin{trivlist}
\item[\hskip \labelsep {\bfseries #1}]}{\end{trivlist}}

\newcommand{\nop}[1]{}

\title{Middleware for LLMs:\\Tools Are Instrumental for Language Agents in Complex Environments}

\author{Yu Gu$^1$, Yiheng Shu$^1$, Hao Yu$^2$, Xiao Liu$^2$, Yuxiao Dong$^2$, Jie Tang$^2$,\\ \textbf{Jayanth Srinivasa$^3$, Hugo Latapie$^3$, Yu Su$^1$}\\
$^1$The Ohio State University \quad $^2$Tsinghua University \quad $^3$Cisco Research\\
 \texttt{\{gu.826, su.809\}@osu.edu}}

\begin{document}
\maketitle
\begin{abstract}
The applications of large language models (LLMs) have expanded well beyond the confines of text processing, signaling a new era where LLMs are envisioned as generalist agents capable of operating within complex environments. 
These environments are often highly expansive, making it impossible for the LLM to process them within its short-term memory.
Motivated by recent research on extending the capabilities of LLMs with tools, 
we seek to investigate the intriguing potential of tools to augment LLMs in handling such complexity by introducing a novel class of tools, termed \textit{middleware}, to aid in the proactive exploration within these massive environments.
Such specialized tools can serve as a middleware layer shielding the LLM from environmental complexity.
In two representative complex environments---knowledge bases (KBs) and databases---we demonstrate the significant potential of augmenting language agents with tools in complex environments.
Notably, equipped with the middleware, GPT-4 achieves \textbf{2.8$\times$} the performance of the best baseline in tasks requiring access to database content and \textbf{2.2$\times$} in KB tasks.
Our findings illuminate the path for advancing language agents in real-world applications.
\footnote{GitHub repo: \href{https://github.com/OSU-NLP-Group/Middleware}{OSU\_NLP/Middleware}\\
\hspace*{1.5em} Hugging Face dataset: \href{https://huggingface.co/datasets/osunlp/KBQA-Agent}{KBQA-Agent}}
\end{abstract}

\section{Introduction}
Large language models (LLMs) have demonstrated a human-like mastery over text~\cite{gpt4,gpt35turbo,llama2,mixtral}.
However, the true ambition of AI extends well beyond the realm of text.
The goal is to ultimately empower LLMs to act as generalist language agents that can aid humans across the multitude of complex real-world tasks~\cite{react, toolformer, agentbench},
which often involve handling complex environments, be it browsing complex webpages~\cite{mind2web}, managing vast databases with millions of entries~\cite{bird}, or querying huge KBs~\cite{pangu}.

\begin{figure}[t]
    \centering
    \includegraphics[width=\linewidth]{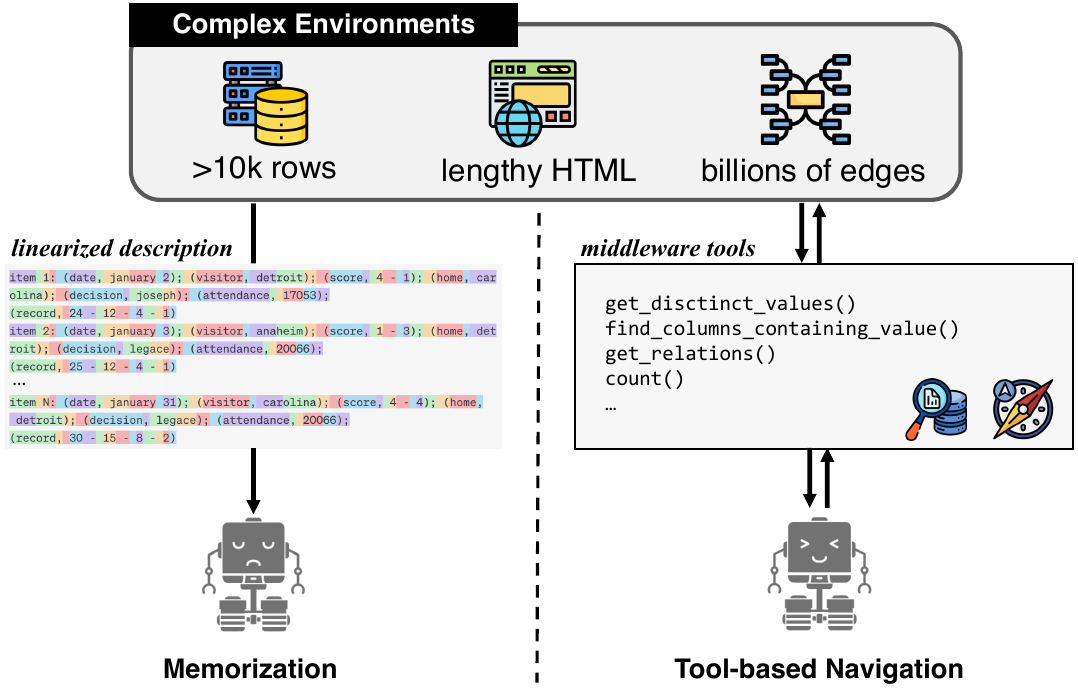}
    \caption{(\textit{left}) When an LLM engages with a complex environment, it can develop an understanding by fitting the environment's description (\ie, linearized tokens) into its short-term memory (\ie, the LLM's input window). 
    However, this method encounters drastic scalability issues as the complexity of the environment grows. (\textit{right}) Another option is to furnish the LLM with a set of tools that assist it in actively engaging with the environment and acquiring the necessary information.
    }
    \label{fig:overview}
\end{figure}

For LLMs to effectively serve as agents that ground human instructions accurately within the environment, they must develop a robust understanding of the environment.
The most direct method to achieve it is to linearize the environment into a sequence of tokens that fit into the LLM's short-term memory (\ie, its input window) and have the LLM process the environment based on the linearized description~\cite{tai2023exploring, alfworld, agentbench}.
However, such a method faces steep challenges in scaling to more complex environments.
Also, discrete token descriptions may not reflect the most natural perception of the environment.
Recent work has explored using tools to extend the boundary of the LLM's capacity~\cite{api-bank, toolllm, toolformer}.
The core idea is that LLMs can actively decide a proper tool to use, using language as a powerful vehicle of thought~\cite{su2023language}. 
Intuitively, we can also equip the LLM with tools that enable navigating complex environments, so that the LLM can proactively invoke different tools to explore the environment, thus circumventing limitations posed by its short-term memory (Figure~\ref{fig:overview}).
However, this promising paradigm has been thus far underexplored.
In this paper, we aim to delve into this paradigm and answer an intriguing question:
\textbf{\textit{How effectively can LLMs handle complex environments with the aid of tools?}}

Answering this question requires equipping the LLM with a suite of tools designed to meet a wide range of needs within the target environment.
In this paper, we carefully develop such tailored tools for two exemplar complex environments, \ie, databases and knowledge bases (KBs).
Unlike readily available Web APIs~\cite{toolllm} used in prior research, our tools have to be manually invented from scratch.
In crafting these tools, we capitalize on the intuition of human information-gathering behaviors—such as performing keyword searches to identify a relevant database column or investigating the connections of a KB entity—to fulfill complex tasks in these environments (Section~\ref{sec:tools}).
Ideally, these tools are designed to function as a \textit{middleware} layer between the LLM and the environment, shielding the LLM from environmental complexity.
With these specialized tools, we propose two novel schemes to enable the LLM to more accurately orchestrate its internal reasoning and tool usage: \textit{error feedback} and \textit{decoupled generation} (Section~\ref{sec:reasoning}).
The combination of the crafted tools and the tool-use schemes allows the LLM to actively explore the environment and ground human instructions into accurate actions.


We evaluate different LLMs on benchmarks featuring complex tasks over the target environments, including a newly curated benchmark for the KB.
The results are revealing: 
\textit{LLMs equipped with customized tools demonstrate a significant enhancement in their ability to engage with complex environments, markedly surpassing the prior art.}
In particular, despite its simplicity, such a middleware layer allows GPT-4~\cite{gpt4} to achieve \textbf{2.8$\times$} the performance (\ie, \num{38.3}\% vs. \num{13.8}\%) of the best baseline in tasks requiring access to database content and \textbf{2.2$\times$} (\ie, \num{59.3}\% vs. \num{27.1}\%) in KB tasks.
Our findings underscore the integral role of tool augmentation in enabling LLMs to handle complex environments.

Our main contributions are as follows: 
a) We develop a new framework with customized tools for two complex environments, to investigate the role of tools in handling complex environments with LLMs;
b) We evaluate six different LLMs on our carefully chosen benchmarks for a comprehensive analysis; 
c) Our analysis highlights a critical takeaway: augmenting LLMs with tools is crucial for successfully tackling complex environments, opening new possibilities to progress LLMs as generalist language agents for practical applications.

\section{Related Work}


\begin{remark} [Interface Complex Environments with LLMs.]
Existing methods that feed the environment directly into the LLM for grounding~\cite{grounding} would fail in complex environments due to scalability issues.
Specifically, these methods process the environment by linearizing it into discrete tokens~\cite{sqlova, alfworld, decaf, agentbench, tai2023exploring, song2023llmplanner}.
However, linearizing expansive environments like databases with millions of entries~\cite{bird} or lengthy webpage HTML code~\cite{mind2web} can often exceed an LLM's input length constraints.
Alternative studies bypass the LLM's direct interaction with complex environments by generating ungrounded draft plans for post-processing grounding~\cite{kb-binder, kb-coder} or by using the LLM to assess grounded plans created via predefined rules~\cite{pangu}.
Such strategies do not fully utilize the LLMs' innate reasoning potential in actively navigating complex environments.
In this paper, we explore a new paradigm where we can bypass these issues by equipping LLMs with a suite of comprehensive tools to actively gather necessary information about the environment upon demand, leveraging the LLMs' inherent reasoning capabilities.
\end{remark}



\begin{remark}[Tool Learning.]
Tools are essential for enhancing the capabilities of LLMs~\cite{toolformer, toollearning, mialon2023augmented, toolkengpt}. 
Existing research, such as ToolLLM~\cite{toolllm} and API-Bank~\cite{api-bank}, focuses on open-domain applications with a wide array of readily available RESTful APIs. 
In contrast, this paper specifically aims to study the potential of tools in augmenting LLMs to effectively execute tasks within complex environments, where we carefully craft the specialized tools for different environments by ourselves.
In addition, research focusing on RESTful APIs typically displays shallow reasoning, while practical tasks within a complex environment typically entail a long sequence of actions (\eg, querying a KB or browsing a webpage).
To enable tool use in more intricate settings within a more specific complex environment, StructGPT~\cite{StructGPT} employs a predefined sequence of tool invocations;
Chameleon~\cite{chameleon} functions in an open-loop setting where the LLM directly produces a sequence for tool usage before any execution occurs.
Both of them fail to seamlessly integrate the reasoning capacity of the LLM with the use of tools.
In this paper, we propose two novel schemes---\textit{error feedback} and \textit{decoupled generation} to more seamlessly and accurately orchestrate the LLM's internal reasoning and tool usage.

\end{remark}



\section{Middleware for LLMs}
We equip LLMs with a suite of tools specifically tailored to support an extensive variety of operations and cater to the diverse needs within a complex environment $\mathcal{E}$.
We call these tools middleware, as they can serve as a feature-rich middle layer between the LLM and $\mathcal{E}$, abstracting the LLM from having to directly interact with all of its intricacies (Section~\ref{sec:tools}).
Furthermore, to fully unleash the inherent reasoning capabilities of LLMs in invoking proper tools, we propose two novel schemes to enhance tool use accuracy: \textit{error feedback}, which provides concrete tool use error information and expects the LLM to correct the error autonomously, and \textit{decoupled generation}, where the LLM's reasoning steps and tool use are separated for better controllability (Section~\ref{sec:reasoning}).
This unified framework allows us to reliably investigate the potential of LLMs in handling complex environments with the aid of tools.

\setlength{\fboxsep}{1pt}
\begin{figure*}[t]
    \centering
    \includegraphics[width=\linewidth]{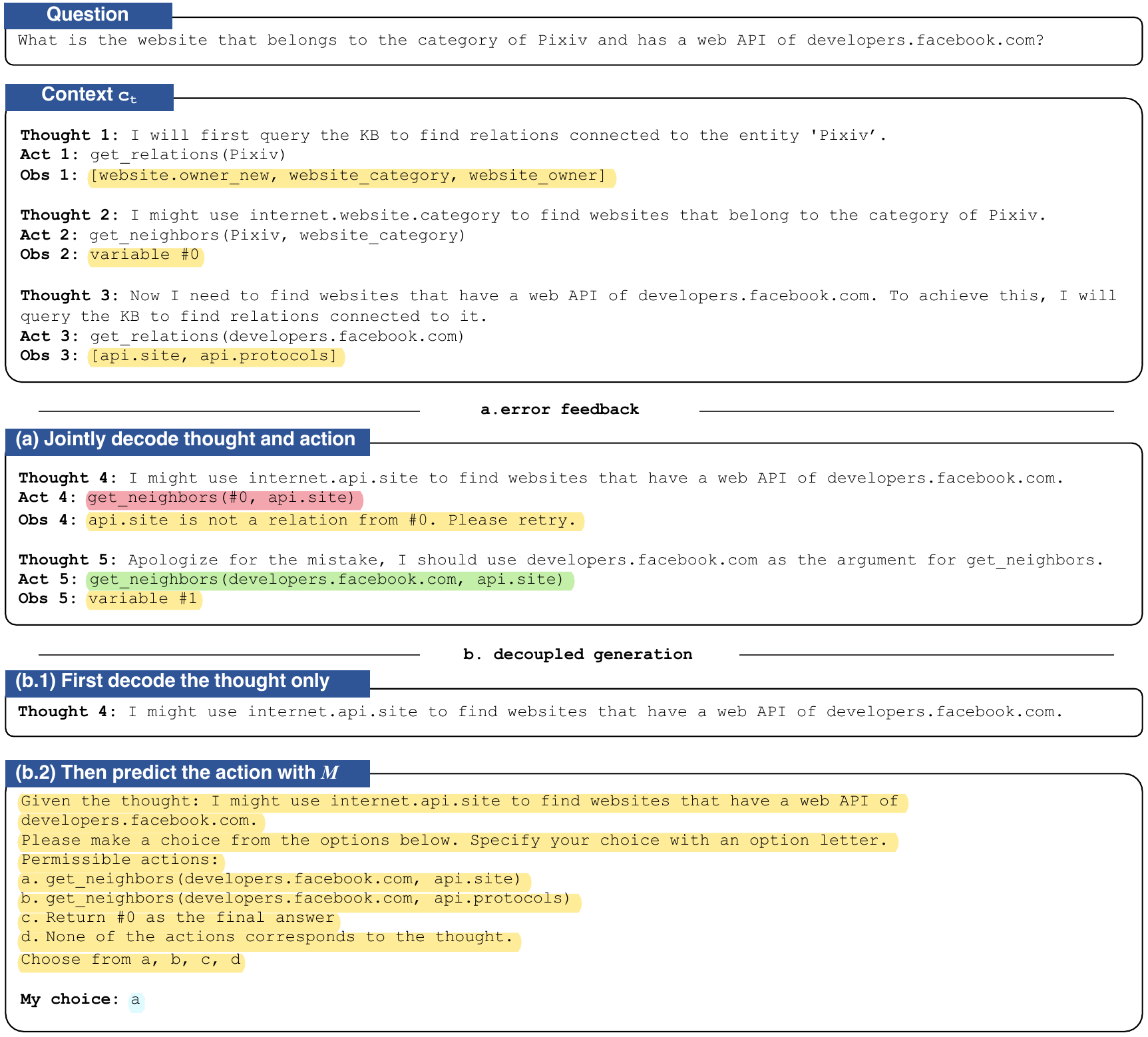}
    \caption{The LLM is equipped with an array of tools to facilitate its engagement with complex environments (\eg, a KB here).
    (a) The LLM may produce invalid actions (marked in \colorbox{mypink}{pink}).
    This can be mitigated by prompting it with an error message that encourages a reattempt (corrected action marked in \colorbox{mygreen}{green}). 
    (b) Alternatively, we can have the LLM first generate a thought, then predict an action based on it in a separate context (marked in \colorbox{myblue}{blue}), and finally insert the action back to the original context.
    Text marked in \colorbox{myyellow}{yellow} are input from the environment.
    }
    \label{fig:reasoning}
\end{figure*}
\subsection{Tools for Complex Environments}
\label{sec:tools}
To evaluate the potential of LLMs in handling complex environments when equipped with tools, we need to first carefully craft the necessary tools for the environments.
These tools should meet two essential criteria:
1) They should offer comprehensiveness, encompassing a broad spectrum of operations and needs.
Broad coverage of tools is crucial for maximizing the potential of LLMs in planning.
2) The tools should prioritize ease of use, enabling the LLM to invoke them mostly with straightforward slot filling, thus shielding the LLM from the implementation details of the tools.

\begin{remark}[Databases]
In production scenarios, databases typically feature dozens of tables, with each table containing thousands of rows or more.
A key task in such environments is performing data analysis through SQL queries.
To bridge the gap between natural language instructions and SQL, LLMs are employed to automate the generation of SQL queries (\ie, text-to-SQL parsing~\cite{spider, bird}).
To support the LLM in crafting complex SQL queries, we introduce a set of specialized tools designed for interaction with intricate databases. 
These tools are divided into two main categories: navigational and functional. 
Navigational tools help the LLM to explore the environment (\eg, \texttt{get\_distinct\_values()} and \texttt{find\_columns\_containing\_value()}),
while functional tools help check each SQL clause composed by the LLM.
For example, \texttt{where()} verifies the legality of the WHERE clause and determines if the specified conditions can match any entries in the database.
In total, we craft \num{12} tools for databases (Appendix~\ref{appendix:db_tools}).
The development of these tools is grounded in our domain expertise in SQL and databases.





% ...
\end{remark}

\begin{remark}[KBs]
Modern KBs, such as \Freebase~\cite{freebase}, are vast repositories storing billions of facts as triples $\langle h, r, t \rangle$.
These KBs cover a wide array of domains and support complex information-seeking tasks, including answering questions that require multi-hop reasoning.
To support the LLM in engaging the extremely massive KB environments, we also devise a toolset tailored for KBs.
Similarly, tools for KBs also include navigational tools and functional tools.
The navigational tools facilitate efficient exploration of the KB by the LLM (\eg, \texttt{get\_relations()} and \texttt{get\_attributes()}), while the functional tools support the LLM in executing precise operations, such as counting and intersecting two sets (\eg, \texttt{intersection()} and \texttt{count()}).
Both are critical for completing complex reasoning tasks on KB.
A key concept in tools for KBs is a \textit{variable}, representing a set of entities and typically generated as an intermediate result through the execution of functions like \texttt{get\_neighbors()} or \texttt{intersection()}.
The use of variables facilitates multi-hop reasoning across KBs, as it enables the natural linkage of a sequence of tool executions.
In total, we implement \num{7} tools for KBs (Appendix~\ref{appendix:kb_tools}).
Our design of KB tools tightly adheres to the common needs in knowledge base question answering (KBQA)~\cite{grailqa, kqapro}.




% ...
\end{remark}


\subsection{Reasoning with Tools}
\label{sec:reasoning}
We first choose ReAct~\cite{react} to serve as the backbone of our reasoning framework, in which the LLM can proactively decide which tool to use based on its own chain-of-thought~\cite{CoT}.
Based on this backbone, we propose two novel schemes to improve the accuracy of tool use, which is critical for complex tasks where successful tool use necessitates careful tool selection and precise argument assignment.
Unlike existing methods relying on human-defined workflows that follow fixed-order tool usage~\cite{StructGPT}, 
our framework allows the LLM autonomy in proactively determining tool selection using CoT.


Formally, at each step $t$, the LLM makes predictions following a policy that maps a current context to an output: $\pi: c_t\rightarrow \hat{a}_t$, where
\begin{align}
    c_t &= (\hat{a}_1, o_1 \cdots, \hat{a}_{t-1}, o_{t-1}) \nonumber\\
    \hat{a}_t &= r_t\oplus a_t \nonumber
\end{align}
$\hat{a}_t$ is the concatenation of a rationale $r_t$ (\ie, a thought in CoT) and a concrete tool use $a_t$ (\eg, in Figure~\ref{fig:reasoning}, $\hat{a}_1$ is the concatenation of \textbf{\texttt{Thought 1}} and \textbf{\texttt{Act 1}}), while $o_t$ is an observation from the environment (\ie, the execution result of $a_t$).
In ReAct, the LLM jointly decodes $\hat{a}_t$ based on $c_t$ for each step.
However, originally designed for simpler tools like the Wikipedia Search API, the naive ReAct framework is more susceptible to producing an invalid $a_t$ that is unfaithful to $r_t$ when applied to more nuanced tool usage.
We propose two simple strategies to remedy this issue.
The first strategy is to simply amplify ReAct by providing detailed \textit{error feedback} in case of incorrect tool usage by the LLM, followed by a prompt to retry based on these messages (see Figure~\ref{fig:reasoning}(a)).\footnote{For databases, we directly use the error message from sqlite3. For KBs, we manually define several simple templates for error feedback along with each tool.}
This relies on the LLM's capacity for self-correction through feedback~\cite{self-correct, self-debug}, which may not always be reliable when the underpinning LLM is weak, potentially leading to the repetition of the same mistakes~\cite{guan2023leveraging}.
Additionally, we present \textit{decoupled generation}, where the LLM's policy $\pi$ is split into two sequential phases (\ie, $\pi\propto \pi_1 \circ \pi_2$), allowing for more nuanced control of its actions.
Initially, the LLM only decodes a thought $r_t$ following $\pi_1(r_t|c_t)$.
Subsequently, the LLM predicts an action $a_t$ in a \textit{separate context}, incorporating both the thought $r_t$ and a set of simple rules $\mathcal{M}$ that determines permissible actions of this step.
This is further guided by $\pi_2$, formulated as $a_t \sim\pi_2(a_t|r_t, \mathcal{M})$.
$\mathcal{M}$ encapsulates the governing rules of the environment (\eg, the relation argument for \texttt{get\_neighbors()} must be derived from the output of \texttt{get\_relations()}, which is applied to the specified entity argument in prior steps), infusing prior knowledge into the LLM's decision-making process (see Figure~\ref{fig:reasoning}(b)).
The concrete prompts used by us are shown in Appendix~\ref{subsec:detailed prompts}.

\begin{table*}[!t]
    \small
    \centering
    \resizebox{0.85\textwidth}{!}{
    \begin{tabular}{lcccccc}
    \toprule
        \textbf{Model} & \multicolumn{2}{c}{\textbf{Req. Cont. (N)}} &  \multicolumn{2}{c}{\textbf{Req. Cont. (Y)} } &  \multicolumn{2}{c}{\textbf{Overall}}\\
        \midrule
        & \textbf{EX}& \textbf{VA}& \textbf{EX}& \textbf{VA}& \textbf{EX}& \textbf{VA}\\
        \midrule
        \rowcolor{lightgrey}
        \multicolumn{7}{c}{\textit{\textbf{w/ Oracle Knowledge}}} \\
        \midrule
        API Docs Prompt~\cite{rajkumar2022api} & & & & & & \\
        \multicolumn{1}{l}{\hspace{35pt} w/ GPT-3.5-turbo}  & \num{38.1}  & \num{78.4} & \num{32.1} & \num{74.6} & \num{36.1} & \num{77.2}\\
        \multicolumn{1}{l}{\hspace{35pt} w/ GPT-4}   & \num{49.5} & \num{95.5} & \num{41.7} & \num{89.9} & \num{46.9} & \num{93.7}\\
        \midrule
        \rowcolor{lightgrey}
        \multicolumn{7}{c}{\textit{\textbf{w/o Oracle Knowledge}}} \\
        \midrule
         API Docs Prompt~\cite{rajkumar2022api} & & & & & & \\
       \multicolumn{1}{l}{\hspace{35pt} w/ GPT-3.5-turbo$^\dag$}  & \num{30.9}  & \num{82.9} & \num{10.9} & \num{80.0} & \num{24.4} & \num{82.0}\\
       \multicolumn{1}{l}{\hspace{35pt} w/ GPT-4}  & \num{38.2}  & \num{91.6} & \num{13.8} & \num{93.1} & \num{30.4} & \num{92.1}\\
        \midrule
        StructGPT~\cite{StructGPT} &  & & & & & \\
        \multicolumn{1}{l}{\hspace{35pt} w/ GPT-3.5-turbo} & \num{36.2} & \num{86.5} & \num{8.7} & \num{80.8} & \num{27.3} & \num{84.7}\\
         \multicolumn{1}{l}{\hspace{35pt} w/ GPT-4} & \num{40.7} & \num{93.4} & \num{13.5} & \num{91.1} & \num{31.8} & \num{92.6}\\
        \midrule
        \OurMethod (\textit{error feedback})  & & & & & &  \\
        \multicolumn{1}{l}{\hspace{35pt} w/ GPT-3.5-turbo} & \textbf{\num{38.8}}  &\textbf{ \num{95.7}} & \textbf{\num{19.8}}  & \textbf{\num{94.7}}  & \textbf{\num{32.7}} & \textbf{\num{95.4}} \\
        \multicolumn{1}{l}{\hspace{35pt} w/ GPT-4} & \textbf{\num{45.1}}  &\textbf{ \num{98.8}} & \textbf{\num{38.3}}  & \textbf{\num{97.2}}  & \textbf{\num{42.9}} & \textbf{\num{98.3}} \\
    \bottomrule
    \end{tabular}
    }
    \caption{Results on \Bird's dev set. Performance of all baselines is obtained under a \textit{zero-shot} setting. $\dag$ denotes the best method \textit{w/o} oracle knowledge on \Bird's official leaderboard. The predictions with API Docs Prompt are directly supplied by the authors of \Bird. }
    \label{table:databases}
\end{table*}
\begin{table*}[!t]
    \small
    \centering
    \resizebox{0.95\textwidth}{!}{
    \begin{tabular}{lcccccccc}
    \toprule
         \textbf{Model} & \multicolumn{2}{c}{\textbf{Counting}} &  \multicolumn{2}{c}{\textbf{Superlative} } &  \multicolumn{2}{c}{\textbf{None}} &  \multicolumn{2}{c}{\textbf{Overall}}\\
         \midrule
          & \textbf{F1}& \textbf{VA}& \textbf{F1}& \textbf{VA}& \textbf{F1}& \textbf{VA} & \textbf{F1}& \textbf{VA}\\
          \midrule
        Pangu$^\diamondsuit$~\cite{pangu} & & & \\
        \multicolumn{1}{l}{\hspace{35pt} w/ GPT-3.5-turbo} &\num{10.1} & \textbf{\num{100.0}} & \num{9.0} & \textbf{\num{100.0}} & \num{23.4} & \textbf{\num{100.0}}& \num{18.1} & \textbf{\num{100.0}} \\
        \multicolumn{1}{l}{\hspace{35pt} w/ GPT-4}   & \num{12.3} &\textbf{\num{100.0}} & \num{14.2} & \textbf{\num{100.0}}& \num{35.6} &\textbf{\num{100.0}} & \num{27.1} & \textbf{\num{100.0}} \\
        \midrule
        KB-Binder~\cite{kb-binder} & & & \\
        \multicolumn{1}{l}{\hspace{35pt} w/ GPT-3.5-turbo (20-shot)} & \num{0.0} & \num{33.7} & \num{0.2} & \num{19.4} & \num{6.7} & \num{37.0} & \num{4.2} & \num{32.8} \\
        \multicolumn{1}{l}{\hspace{35pt} w/ GPT-4 (20-shot)}   & \num{7.9} & \num{48.3} & \num{0.4} & \num{28.2} & \num{6.0} & \num{45.8} & \num{5.2} & \num{42.6} \\
        \midrule
        StructGPT~\cite{StructGPT} & & & \\
        \multicolumn{1}{l}{\hspace{35pt} w/ GPT-3.5-turbo} & \num{4.5} & \num{50.6} & \num{3.9} & \num{51.5} & \num{11.4} & \num{57.1} & \num{8.6} & \num{54.8} \\
        \multicolumn{1}{l}{\hspace{35pt} w/ GPT-4}   & \num{2.2} & \num{37.1} & \num{3.9} & \num{30.1} & \num{11.7} & \num{26.3} & \num{8.4} & \num{29.0} \\
        \midrule
        \OurMethod (\textit{error feedback}) & & & \\
        \multicolumn{1}{l}{\hspace{35pt} w/ GPT-3.5-turbo} & \num{33.7} & \num{70.7} & \num{22.0} & \num{64.1} & \num{23.9} & \num{56.8} & \num{25.3} &  \num{60.8} \\
        \multicolumn{1}{l}{\hspace{35pt} w/ GPT-4} & \num{70.7} & \num{96.6} & \num{39.9} & \num{74.5} & \num{55.8} & \num{74.0} & \num{55.1} & \num{78.0} \\
        \midrule
        \OurMethod (\textit{decoupled generation}) & & & \\
        \multicolumn{1}{l}{\hspace{35pt} w/ GPT-3.5-turbo} & \textbf{\num{48.9}} & \num{97.7} & \textbf{\num{29.5}} & \num{88.0} & \textbf{\num{32.1}} & \num{77.3} & \textbf{\num{34.3}} & \num{83.0} \\
        \multicolumn{1}{l}{\hspace{35pt} w/ GPT-4} & \textbf{\num{74.1}} & \num{98.9} & \textbf{\num{42.6}} & \num{85.1} & \textbf{\num{61.0}} & \num{83.6} & \textbf{\num{59.3}} & \num{85.8} \\
    \bottomrule
    \end{tabular}}
    \caption{Results on \KBQAAgent. All models are provided with \textit{one-shot} demonstration except for KB-Binder, where we provide 20-shot demonstrations for optimal performance. $\diamondsuit$ indicates our reimplementation of Pangu, as the original code lacks support for chat models. We assume perfect entity linking for all methods.} 
    \label{table:kb}
\end{table*}

\section{Benchmarks}\label{sec:benchmark}

The predominant tasks for databases and KBs are text-to-SQL parsing and KBQA.
However, \textit{popular benchmarks for them may fall short for evaluating language agents out-of-box.}
Specifically, the majority of questions in popular KBQA datasets like \WebQSP~\cite{berant-etal-2013-semantic, yih-etal-2016-value} are one-hop or two-hop questions, for which we can effectively handle with existing semantic parsing methods~\cite{gu2022knowledge}.
Additionally, the databases featured in \Spider~\cite{spider} and \WikiSQL~\cite{wikisql} have limited complexity in terms of both schema design and the number of rows in the tables.
This over-simplification enables the direct feeding of the database schema to the LLM, achieving strong performance without the need to access the actual content of the database~\cite{rajkumar2022api}.
Therefore, we need different benchmarks with complex environments and instructions that better mirror the real-world situations language agents must handle (see statistics of our benchmarks in Appendix~\ref{appendix:benchmarks}).

\begin{remark} [Databases]
For databases, we leverage \Bird~\cite{bird}, which is a recent dataset notable for its complexity, featuring intricate instructions over highly complex databases.
There are originally two different settings in \Bird: with and without oracle knowledge,
where the oracle knowledge supplies specific information about the target database needed to fulfill each task.
For instance, \textit{``{Exclusively virtual refers to Virtual = `F'}''}.
With such oracle knowledge, the complexity of the environments is substantially mitigated; 
it offers a shortcut for the task and eliminates the necessity for deep engagement with the database.
This cheating setting is also unrealistic for practical applications.
As a result, we stick to the setting without oracle knowledge.
For each of the \num{1534} questions in \Bird's dev set, we manually label whether accessing the database content is necessary to compile the SQL queries, noting that access is unnecessary if all mentioned values in a question exactly match database cells.
This facilitates decomposing the language agent's performance based on questions that require deeper database engagement (\num{496} questions) versus not (\num{1038} questions) and enables fine-grained insights into the LLM's performance.
In addition to execution accuracy (\textbf{EX}) used in \Bird, which determines if the execution results of the predicted SQL match those of the ground truth SQL,
we also evaluate whether the predicted SQL is a valid SQL query (\textbf{VA}).
\end{remark}

\begin{remark} [KBs]
We curate \KBQAAgent, a new test set sourcing from existing KBQA datasets that contain complex questions.
In particular, we selected \num{500} diverse questions that involve at least three relations, or two relations coupled with an aggregation function (\ie, \textbf{Counting} or \textbf{Superlative}).
For each question, we annotate it with a ground truth sequence of actions based on the toolset defined by us.\footnote{We leverage the gold S-expressions provided by~\citet{gu-su-2022-arcaneqa}. 
}
Specifically, \KBQAAgent comprises questions from three KBQA datasets on \Freebase: \GrailQ~\cite{grailqa}, \ComplexQ~\cite{cwq}, and \GraphQ~\cite{graphq},
ensuring a wide range of question types and sources.
\KBQAAgent is designed to be more representative of challenging, real-world scenarios compared to existing benchmarks (Appendix~\ref{appendix:benchmarks}).
It offers an ideal testbed for evaluating language agents in interacting with massive KBs.
We assess this through two metrics:
\textbf{F1} of answer entities and Validity (\textbf{VA}), a binary metric evaluating the LLM's ability to find an answer, whether correct or not.
\end{remark}

\begin{figure*}
    \centering
    \includegraphics[width=0.9\linewidth]{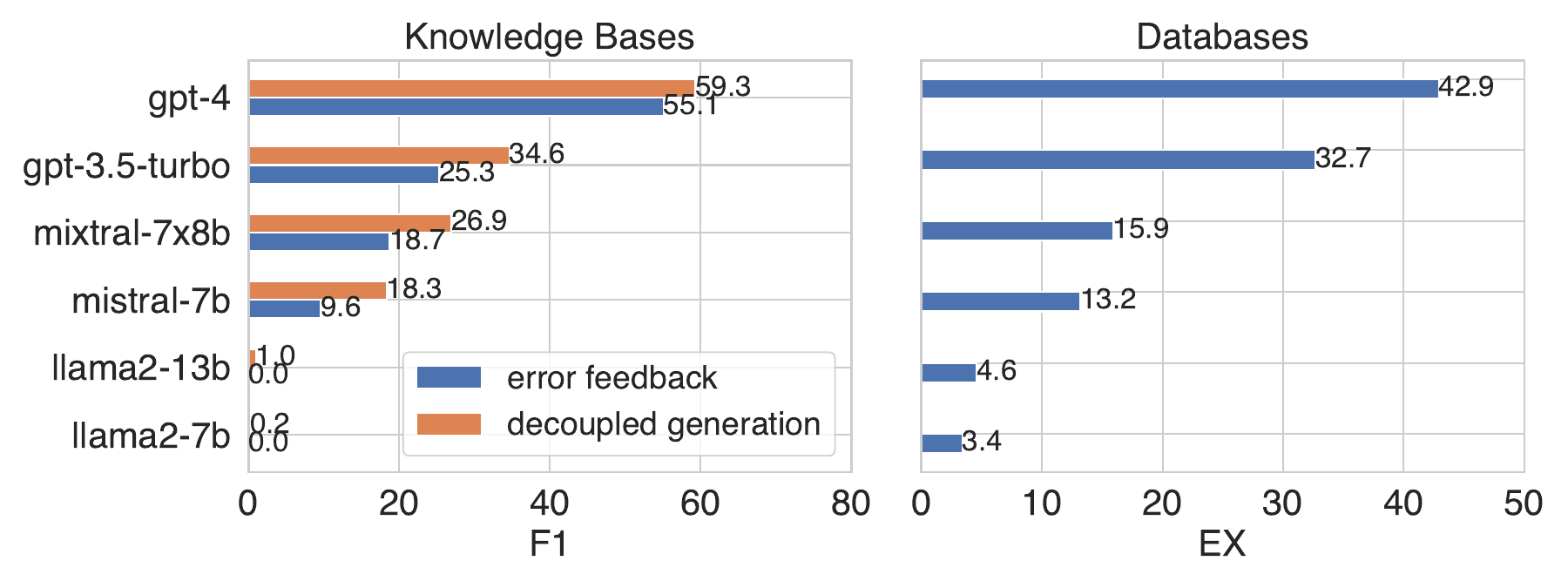}
    \caption{The open-source LLMs still largely lag behind GPT-3.5-turbo and GPT-4 in both environments.\vspace{-10pt}}
    \label{fig:hist}
\end{figure*}
\section{Experiments}
\subsection{Setup}
\begin{remark}[Implementation]
To concretely instantiate our tools for the two environments, we employ standard query interfaces for databases and KBs, specifically SQLite for databases and Virtuoso for KBs.
We then prompt the LLM with the tool descriptions together with the input task instructions (Appendix~\ref{subsec:detailed prompts}).
Each environment exhibits its own unique characteristics and challenges.
In KBQA, the arguments for each function are either a variable or an item from the KB schema (\ie, a relation or an attribute). 
In contrast, in text-to-SQL parsing, the arguments can be more varied, ranging from a part of a SQL query to a complete query. 
This makes listing potential actions, as needed in \textit{decoupled generation}, much more complex for text-to-SQL parsing.
Therefore, we implement \textit{error feedback} solely for text-to-SQL parsing. 

For the underlying LLMs, we primarily compare \OurMethod with baseline methods using two of the most advanced LLMs---GPT-3.5-turbo-0613~\cite{gpt35turbo} and GPT-4-0613~\cite{gpt4}---since our goal is investigating the full potential of tool-enhanced LLMs operating within complex environments.
In addition, we also explore four open-source LLMs to more comprehensively evaluate our framework:  Llama2-7B-Chat, Llama2-13B-Chat~\cite{llama2}, Mistral-7B-Instruct-v0.2~\cite{mistral}, and Mixtral 8$\times$7B-Instruct-v0.1~\cite{mixtral}.
\end{remark}
\begin{figure}
    \centering
    \includegraphics[width=\linewidth]{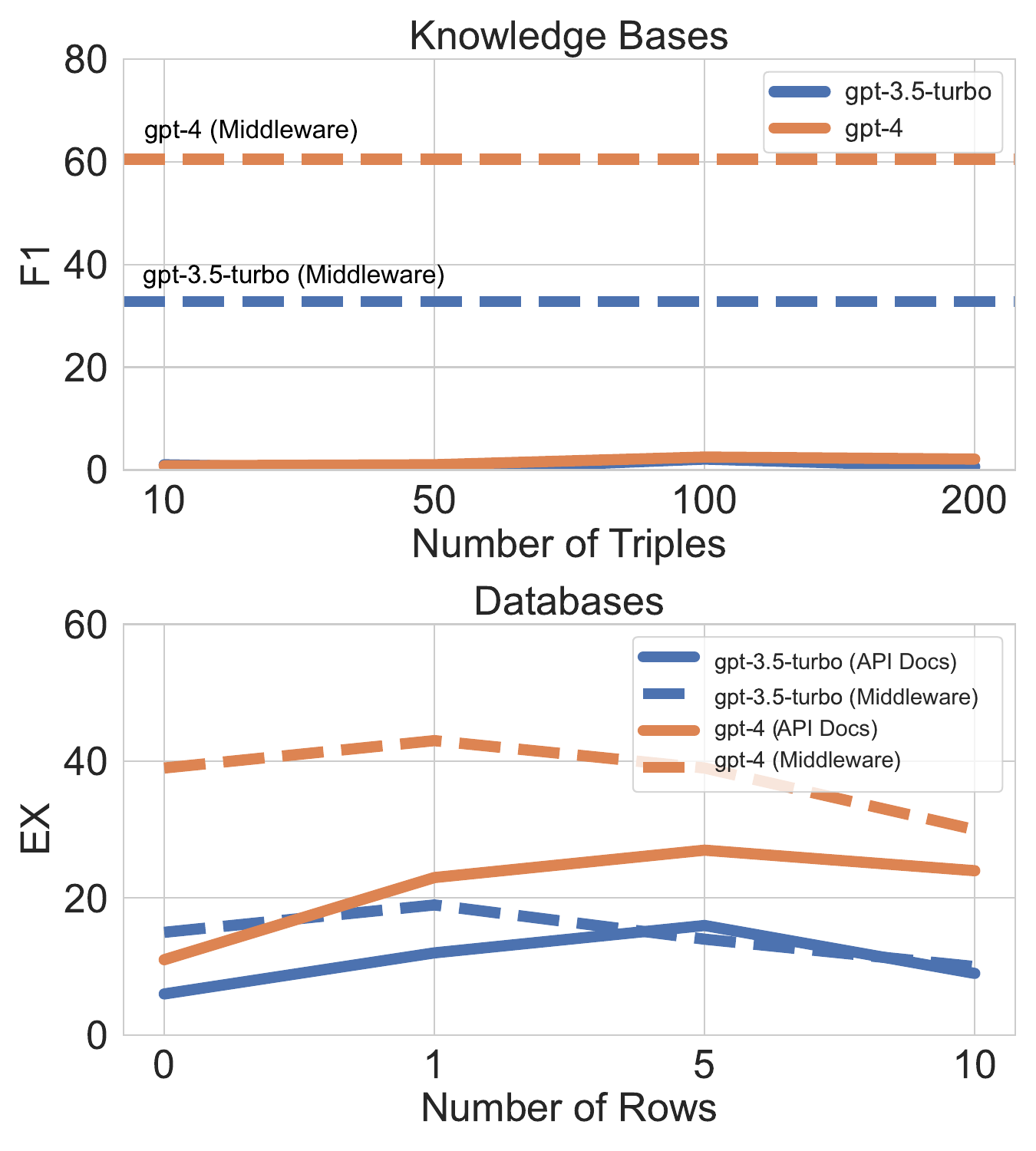}
    \caption{The customized tools can serve as effective \textit{middleware} between the LLM and the environment.\vspace{-15pt}}
    \label{fig:line}
\end{figure}

\vspace{-5pt}
\begin{remark}[Baselines]
To fully understand the potential of tool augmentation for assisting LLMs in handling complex environments, we compare \OurMethod against an array of strong baselines.
For text-to-SQL parsing, LLMs demonstrate a strong ability to compose SQL queries when properly prompted with the database schema (\ie, API docs prompting~\cite{rajkumar2022api}).
This also represents the current state-of-the-art prompting-based method when oracle knowledge is not available on \Bird's leaderboard. 
In adddition, we also compare with more baselines on \Bird's leaderboard that originally did not submit their results using no oracle knowledge (See Appendix~\ref{appendix:exp_comp}).
For all methods on text-to-SQL parsing, we adopt the \textit{zero-shot} setting.
Unlike text-to-SQL parsing, directly prompting LLMs does not generate reasonable outputs for KBQA due to the massive size of the KB schema.
Instead, existing KBQA methods based on LLMs typically follow two paradigms: either first generating an ungrounded program and then grounding the program to the KB schema afterwards~\cite{kb-binder, kb-coder}, or gradually constructing a complex program and grounding it step by step~\cite{pangu}.
We compare \OurMethod with the most representative work from each paradigm, namely KB-Binder~\cite{kb-binder} and Pangu~\cite{pangu}.
We also include StructGPT as an additional baseline for tool use.
For all KBQA methods except KB-Binder, we provide a \textit{one-shot} demo to obtain more meaningful results.
\end{remark}

\subsection{Main Results}
As shown in Tables~\ref{table:databases} and \ref{table:kb}, equipping LLMs with customized tools leads to significant improvement over previous standards, almost doubling or tripling the performance under multiple metrics.
Specifically, API docs prompting can only feed the schema information to the LLM due to the vast amount of database content.
As a result, it fails catastrophically on examples that require database content to compose the SQL query.
In contrast, \OurMethod equips the agent with tools to actively navigate the database to collect relevant information for composing a SQL query.
As a result, \OurMethod significantly closes the gap between performance on questions requiring database content and questions not requiring it when using GPT-4 (\ie, \num{45.1}\% vs. \num{38.3}\%).
Additionally, we notice that \OurMethod minimizes the gap between with and without oracle knowledge from \num{15.5}\% to \num{4.0}\% using GPT-4 and \num{11.7}\% to \num{3.3}\% using GPT-3.5-turbo.
Finally, StructGPT demonstrates a similar trend to API docs prompting because its tools do not provide any information about the database content.
For KBQA, \OurMethod demonstrates uniformly superior performance across different question types and significantly outperforms Pangu with both GPT-3.5-turbo and GPT-4.
In particular, when equipped with GPT-4, \OurMethod + \textit{decoupled generation} outperforms Pangu by \num{32.2}\% in F1.
As for the other two baselines, KB-Binder and StructGPT, both fail miserably on our challenging setting.
On the one hand, KB-Binder only retrieves relations within two hops from the entities for grounding. 
However, most questions in \KBQAAgent involve more than two relations.
As a result, many of its drafted programs are unable to ground, which explains its low VA.
On the other hand, StructGPT is heavily limited by its constrained toolset \nop{(\ie, only \texttt{extract\_neighbor\_relations} and \texttt{extract\_triples})} and cannot handle complex questions in \KBQAAgent.
Therefore, StructGPT frequently refuses to provide an answer (as revealed by its low VA) due to insufficient information.
The strong performance of \OurMethod underscores that tools are instrumental for language agents in complex environments.

Due to the space limit, we provide additional results in Appendix~\ref{appendix:exp}.
\subsection{Experiments with Open-Source LLMs}
To gain a more thorough insight, we also include experiments with four open-source LLMs ( Figure~\ref{fig:hist}). 
Our findings indicate that Llama2 models generally underperform compared to other LLMs, aligning with trends observed in other LLM leaderboards, such as Chatbot Arena~\cite{chatbot_arena}.
Specifically, we find Llama2 models struggle with even generating grammatical tool use following our instruction.
On the other hand, Mistral and Mixtral demonstrate much better performance than Llama2.
In particular, Mixtral represents an advanced mixture-of-experts model that has demonstrated superior performance and even surpasses GPT-3.5-turbo on Chatbot Arena~\cite{chatbot_arena}.
However, different from answering open-ended questions, properly engaging with the complex environment demands the language agent to produce more precise actions that strictly conform to the task specification.
There is still a gap between Mixtral and GPT-3.5-turbo in terms of predicting valid actions over complex environments.
Compared to GPT-3.5-turbo, Mixtral tends to output invalid actions more frequently.
This also explains why \textit{decoupled generation}, where the output space is strictly constrained to a list of valid actions, helps weaker models more.
With \OurMethod + \textit{decoupled generation}, using Mistral can almost match the best baseline performance with GPT-3.5-turbo, and using Mixtral can even match the best baseline with GPT-4.
While stronger models like GPT-4 can effectively recover the mistake via \textit{error feedback},
weaker models tend to benefit more from \textit{decoupled generation}.

\subsection{Tools as A Middleware Layer}
\label{sec:exp_middleware}
To deepen our understanding of the integral roles of tools in aiding LLMs in accessing complex environments (\ie, KB triples and database rows in our setup), we conduct further analysis by comparing \OurMethod with prompting baselines with different amounts of data items directly sampled from the environment (Figure~\ref{fig:line}).
For the KB, we sample \num{10}, \num{50}, \num{100}, and \num{200} triples from \Freebase based on the three-hop neighborhood of each entity in a question.
These triples are the top-ranked ones using a sentence-BERT retriever~\cite{reimers-gurevych-2019-sentence} based on their similarity with the input question.
We prompt the LLM directly with these sampled triples and request it to generate an answer to the given question.
Given the extensive size of \Freebase, accurately representing the environment with a mere subset of samples proves to be exceedingly difficult. Consequently, both GPT-3.5 Turbo and GPT-4 consistently yield an F1 score close to \num{0}. 
For the database, we similarly augment API docs prompting with $1$, $5$, and $10$ sampled rows for each table and evaluate on \num{100} random questions from \Bird that require accessing database content.
Additionally, we also augment \OurMethod with the same sampled rows in the database setting.
We observe that including more database rows initially boosts baseline performance but eventually decreases it.
With \OurMethod, prompting the LLM with sampled rows yields minimal gain, and the standard setting without sampled rows already significantly outperforms all baselines.
These results further confirm that the LLM, when augmented with tools, can effectively engage with complex environments, flexibly gathering the necessary information on demand and bypassing the limitations on the amount of data it can handle (\eg, around \num{200} triples or \num{10} rows per table).




\section{Conclusion}
A pioneering vision is for language agents to assist humans in tackling intricate real-world tasks.
This paper demonstrates that with meticulously-crafted tools acting as \textit{middleware} between LLMs and complex environments, LLMs can substantially exceed current solutions.
Our results spotlight these specialized tools’ indispensable role in unlocking the potential of LLMs within complex real-world tasks previously posing immense challenges.
\section*{Limitations}
In this paper, we aim to address the compelling question we posed: how effectively can LLMs handle complex environments with the aid of tools? 
We investigate this through evaluations in two exemplary environments: KBs and databases. 
While we achieve notable results in these environments, it is important to acknowledge that implementing customized tools for KBs and databases presents fewer challenges compared to environments without a straightforward query interface, such as a webpage or a physical environment.
In future work, we plan to extend \OurMethod across a broader range of environments, aiming to fully realize the potential of language agents in complex environments through the integration of customized middleware tools.

Furthermore, the tools developed in this study are soley grounded in our experience. 
Despite this, our results already demonstrate the significant potential of augmenting LLMs with customized tools in complex environments, aligning with the primary objective of this paper. 
Nonetheless, to enhance performance further, adopting a more principled strategy for tool design is essential.
Additionally, investigating autonomous tool-making methods~\cite{trove} in complex environments presents a promising direction for future research.


\bibliography{anthology,custom}

\begin{thebibliography}{50}
\expandafter\ifx\csname natexlab\endcsname\relax\def\natexlab#1{#1}\fi

\bibitem[{Berant et~al.(2013)Berant, Chou, Frostig, and Liang}]{berant-etal-2013-semantic}
Jonathan Berant, Andrew Chou, Roy Frostig, and Percy Liang. 2013.
\newblock \href {https://aclanthology.org/D13-1160} {Semantic parsing on {F}reebase from question-answer pairs}.
\newblock In \emph{Proceedings of the 2013 Conference on Empirical Methods in Natural Language Processing}, pages 1533--1544, Seattle, Washington, USA. Association for Computational Linguistics.

\bibitem[{Bollacker et~al.(2008)Bollacker, Evans, Paritosh, Sturge, and Taylor}]{freebase}
Kurt~D. Bollacker, Colin Evans, Praveen~K. Paritosh, Tim Sturge, and Jamie Taylor. 2008.
\newblock \href {https://doi.org/10.1145/1376616.1376746} {Freebase: a collaboratively created graph database for structuring human knowledge}.
\newblock In \emph{Proceedings of the {ACM} {SIGMOD} International Conference on Management of Data, {SIGMOD} 2008, Vancouver, BC, Canada, June 10-12, 2008}, pages 1247--1250. {ACM}.

\bibitem[{Cao et~al.(2022)Cao, Shi, Pan, Nie, Xiang, Hou, Li, He, and Zhang}]{kqapro}
Shulin Cao, Jiaxin Shi, Liangming Pan, Lunyiu Nie, Yutong Xiang, Lei Hou, Juanzi Li, Bin He, and Hanwang Zhang. 2022.
\newblock \href {https://doi.org/10.18653/v1/2022.acl-long.422} {{KQA Pro}: {A} dataset with explicit compositional programs for complex question answering over knowledge base}.
\newblock In \emph{Proceedings of the 60th Annual Meeting of the Association for Computational Linguistics (Volume 1: Long Papers), {ACL} 2022, Dublin, Ireland, May 22-27, 2022}, pages 6101--6119. Association for Computational Linguistics.

\bibitem[{Chandu et~al.(2021)Chandu, Bisk, and Black}]{grounding}
Khyathi~Raghavi Chandu, Yonatan Bisk, and Alan~W. Black. 2021.
\newblock \href {https://doi.org/10.18653/V1/2021.FINDINGS-ACL.375} {Grounding 'grounding' in {NLP}}.
\newblock In \emph{Findings of the Association for Computational Linguistics: {ACL/IJCNLP} 2021, Online Event, August 1-6, 2021}, volume {ACL/IJCNLP} 2021 of \emph{Findings of {ACL}}, pages 4283--4305. Association for Computational Linguistics.

\bibitem[{Chen et~al.(2023)Chen, Lin, Sch{\"{a}}rli, and Zhou}]{self-debug}
Xinyun Chen, Maxwell Lin, Nathanael Sch{\"{a}}rli, and Denny Zhou. 2023.
\newblock \href {https://doi.org/10.48550/ARXIV.2304.05128} {Teaching large language models to self-debug}.
\newblock \emph{CoRR}, abs/2304.05128.

\bibitem[{Deng et~al.(2023)Deng, Gu, Zheng, Chen, Stevens, Wang, Sun, and Su}]{mind2web}
Xiang Deng, Yu~Gu, Boyuan Zheng, Shijie Chen, Samuel Stevens, Boshi Wang, Huan Sun, and Yu~Su. 2023.
\newblock \href {https://doi.org/10.48550/ARXIV.2306.06070} {Mind2web: Towards a generalist agent for the web}.
\newblock \emph{CoRR}, abs/2306.06070.

\bibitem[{Gao et~al.(2024)Gao, Wang, Li, Sun, Qian, Ding, and Zhou}]{dial-sql}
Dawei Gao, Haibin Wang, Yaliang Li, Xiuyu Sun, Yichen Qian, Bolin Ding, and Jingren Zhou. 2024.
\newblock \href {https://www.vldb.org/pvldb/vol17/p1132-gao.pdf} {Text-to-sql empowered by large language models: {A} benchmark evaluation}.
\newblock \emph{Proc. {VLDB} Endow.}, 17(5):1132--1145.

\bibitem[{Gou et~al.(2023)Gou, Shao, Gong, Shen, Yang, Duan, and Chen}]{self-correct}
Zhibin Gou, Zhihong Shao, Yeyun Gong, Yelong Shen, Yujiu Yang, Nan Duan, and Weizhu Chen. 2023.
\newblock \href {https://doi.org/10.48550/ARXIV.2305.11738} {{CRITIC:} large language models can self-correct with tool-interactive critiquing}.
\newblock \emph{CoRR}, abs/2305.11738.

\bibitem[{Gu et~al.(2023)Gu, Deng, and Su}]{pangu}
Yu~Gu, Xiang Deng, and Yu~Su. 2023.
\newblock \href {https://doi.org/10.18653/v1/2023.acl-long.270} {Don't generate, discriminate: {A} proposal for grounding language models to real-world environments}.
\newblock In \emph{Proceedings of the 61st Annual Meeting of the Association for Computational Linguistics (Volume 1: Long Papers), {ACL} 2023, Toronto, Canada, July 9-14, 2023}, pages 4928--4949. Association for Computational Linguistics.

\bibitem[{Gu et~al.(2021)Gu, Kase, Vanni, Sadler, Liang, Yan, and Su}]{grailqa}
Yu~Gu, Sue Kase, Michelle Vanni, Brian~M. Sadler, Percy Liang, Xifeng Yan, and Yu~Su. 2021.
\newblock \href {https://doi.org/10.1145/3442381.3449992} {Beyond {I.I.D.:} three levels of generalization for question answering on knowledge bases}.
\newblock In \emph{{WWW} '21: The Web Conference 2021, Virtual Event / Ljubljana, Slovenia, April 19-23, 2021}, pages 3477--3488. {ACM} / {IW3C2}.

\bibitem[{Gu et~al.(2022)Gu, Pahuja, Cheng, and Su}]{gu2022knowledge}
Yu~Gu, Vardaan Pahuja, Gong Cheng, and Yu~Su. 2022.
\newblock \href {https://doi.org/10.48550/arXiv.2209.04994} {Knowledge base question answering: A semantic parsing perspective}.
\newblock In \emph{4th Conference on Automated Knowledge Base Construction}.

\bibitem[{Gu and Su(2022)}]{gu-su-2022-arcaneqa}
Yu~Gu and Yu~Su. 2022.
\newblock \href {https://aclanthology.org/2022.coling-1.148} {{A}rcane{QA}: Dynamic program induction and contextualized encoding for knowledge base question answering}.
\newblock In \emph{Proceedings of the 29th International Conference on Computational Linguistics}, pages 1718--1731, Gyeongju, Republic of Korea. International Committee on Computational Linguistics.

\bibitem[{Guan et~al.(2023)Guan, Valmeekam, Sreedharan, and Kambhampati}]{guan2023leveraging}
Lin Guan, Karthik Valmeekam, Sarath Sreedharan, and Subbarao Kambhampati. 2023.
\newblock \href {https://doi.org/10.48550/ARXIV.2305.14909} {Leveraging pre-trained large language models to construct and utilize world models for model-based task planning}.
\newblock \emph{CoRR}, abs/2305.14909.

\bibitem[{Hao et~al.(2023)Hao, Liu, Wang, and Hu}]{toolkengpt}
Shibo Hao, Tianyang Liu, Zhen Wang, and Zhiting Hu. 2023.
\newblock \href {https://doi.org/10.48550/arXiv.2305.11554} {Toolkengpt: Augmenting frozen language models with massive tools via tool embeddings}.
\newblock \emph{CoRR}, abs/2305.11554.

\bibitem[{Hwang et~al.(2019)Hwang, Yim, Park, and Seo}]{sqlova}
Wonseok Hwang, Jinyeung Yim, Seunghyun Park, and Minjoon Seo. 2019.
\newblock \href {http://arxiv.org/abs/1902.01069} {A comprehensive exploration on wikisql with table-aware word contextualization}.
\newblock \emph{CoRR}, abs/1902.01069.

\bibitem[{Jiang et~al.(2023{\natexlab{a}})Jiang, Sablayrolles, Mensch, Bamford, Chaplot, de~Las~Casas, Bressand, Lengyel, Lample, Saulnier, Lavaud, Lachaux, Stock, Scao, Lavril, Wang, Lacroix, and Sayed}]{mistral}
Albert~Q. Jiang, Alexandre Sablayrolles, Arthur Mensch, Chris Bamford, Devendra~Singh Chaplot, Diego de~Las~Casas, Florian Bressand, Gianna Lengyel, Guillaume Lample, Lucile Saulnier, L{\'{e}}lio~Renard Lavaud, Marie{-}Anne Lachaux, Pierre Stock, Teven~Le Scao, Thibaut Lavril, Thomas Wang, Timoth{\'{e}}e Lacroix, and William~El Sayed. 2023{\natexlab{a}}.
\newblock \href {https://doi.org/10.48550/ARXIV.2310.06825} {Mistral 7b}.
\newblock \emph{CoRR}, abs/2310.06825.

\bibitem[{Jiang et~al.(2024)Jiang, Sablayrolles, Roux, Mensch, Savary, Bamford, Chaplot, de~Las~Casas, Hanna, Bressand, Lengyel, Bour, Lample, Lavaud, Saulnier, Lachaux, Stock, Subramanian, Yang, Antoniak, Scao, Gervet, Lavril, Wang, Lacroix, and Sayed}]{mixtral}
Albert~Q. Jiang, Alexandre Sablayrolles, Antoine Roux, Arthur Mensch, Blanche Savary, Chris Bamford, Devendra~Singh Chaplot, Diego de~Las~Casas, Emma~Bou Hanna, Florian Bressand, Gianna Lengyel, Guillaume Bour, Guillaume Lample, L'elio~Renard Lavaud, Lucile Saulnier, Marie-Anne Lachaux, Pierre Stock, Sandeep Subramanian, Sophia Yang, Szymon Antoniak, Teven~Le Scao, Th{\'e}ophile Gervet, Thibaut Lavril, Thomas Wang, Timoth{\'e}e Lacroix, and William~El Sayed. 2024.
\newblock \href {http://arxiv.org/abs/2401.04088} {Mixtral of experts}.
\newblock \emph{CoRR}.

\bibitem[{Jiang et~al.(2023{\natexlab{b}})Jiang, Zhou, Dong, Ye, Zhao, and Wen}]{StructGPT}
Jinhao Jiang, Kun Zhou, Zican Dong, Keming Ye, Wayne~Xin Zhao, and Ji{-}Rong Wen. 2023{\natexlab{b}}.
\newblock \href {https://doi.org/10.48550/arXiv.2305.09645} {{StructGPT}: {A} general framework for large language model to reason over structured data}.
\newblock \emph{CoRR}, abs/2305.09645.

\bibitem[{Li et~al.(2023{\natexlab{a}})Li, Hui, Qu, Li, Yang, Li, Wang, Qin, Cao, Geng, Huo, Zhou, Ma, Li, Chang, Huang, Cheng, and Li}]{bird}
Jinyang Li, Binyuan Hui, Ge~Qu, Binhua Li, Jiaxi Yang, Bowen Li, Bailin Wang, Bowen Qin, Rongyu Cao, Ruiying Geng, Nan Huo, Xuanhe Zhou, Chenhao Ma, Guoliang Li, Kevin~Chen{-}Chuan Chang, Fei Huang, Reynold Cheng, and Yongbin Li. 2023{\natexlab{a}}.
\newblock \href {https://doi.org/10.48550/ARXIV.2305.03111} {Can {LLM} already serve as {A} database interface? {A} big bench for large-scale database grounded text-to-sqls}.
\newblock \emph{CoRR}, abs/2305.03111.

\bibitem[{Li et~al.(2023{\natexlab{b}})Li, Song, Yu, Yu, Li, Huang, and Li}]{api-bank}
Minghao Li, Feifan Song, Bowen Yu, Haiyang Yu, Zhoujun Li, Fei Huang, and Yongbin Li. 2023{\natexlab{b}}.
\newblock \href {https://doi.org/10.48550/arXiv.2304.08244} {{API-Bank}: {A} benchmark for tool-augmented llms}.
\newblock \emph{CoRR}, abs/2304.08244.

\bibitem[{Li et~al.(2023{\natexlab{c}})Li, Ma, Zhuang, Gu, Su, and Chen}]{kb-binder}
Tianle Li, Xueguang Ma, Alex Zhuang, Yu~Gu, Yu~Su, and Wenhu Chen. 2023{\natexlab{c}}.
\newblock \href {https://api.semanticscholar.org/CorpusID:258461017} {Few-shot in-context learning on knowledge base question answering}.
\newblock In \emph{Annual Meeting of the Association for Computational Linguistics}.

\bibitem[{Liu et~al.(2023)Liu, Yu, Zhang, Xu, Lei, Lai, Gu, Ding, Men, Yang, Zhang, Deng, Zeng, Du, Zhang, Shen, Zhang, Su, Sun, Huang, Dong, and Tang}]{agentbench}
Xiao Liu, Hao Yu, Hanchen Zhang, Yifan Xu, Xuanyu Lei, Hanyu Lai, Yu~Gu, Hangliang Ding, Kaiwen Men, Kejuan Yang, Shudan Zhang, Xiang Deng, Aohan Zeng, Zhengxiao Du, Chenhui Zhang, Sheng Shen, Tianjun Zhang, Yu~Su, Huan Sun, Minlie Huang, Yuxiao Dong, and Jie Tang. 2023.
\newblock \href {https://doi.org/10.48550/arXiv.2308.03688} {{AgentBench}: Evaluating llms as agents}.
\newblock \emph{CoRR}, abs/2308.03688.

\bibitem[{Lu et~al.(2023)Lu, Peng, Cheng, Galley, Chang, Wu, Zhu, and Gao}]{chameleon}
Pan Lu, Baolin Peng, Hao Cheng, Michel Galley, Kai{-}Wei Chang, Ying~Nian Wu, Song{-}Chun Zhu, and Jianfeng Gao. 2023.
\newblock \href {https://doi.org/10.48550/ARXIV.2304.09842} {Chameleon: Plug-and-play compositional reasoning with large language models}.
\newblock \emph{CoRR}, abs/2304.09842.

\bibitem[{Mialon et~al.(2023)Mialon, Dess{\`{\i}}, Lomeli, Nalmpantis, Pasunuru, Raileanu, Rozi{\`{e}}re, Schick, Dwivedi{-}Yu, Celikyilmaz, Grave, LeCun, and Scialom}]{mialon2023augmented}
Gr{\'{e}}goire Mialon, Roberto Dess{\`{\i}}, Maria Lomeli, Christoforos Nalmpantis, Ramakanth Pasunuru, Roberta Raileanu, Baptiste Rozi{\`{e}}re, Timo Schick, Jane Dwivedi{-}Yu, Asli Celikyilmaz, Edouard Grave, Yann LeCun, and Thomas Scialom. 2023.
\newblock \href {https://doi.org/10.48550/ARXIV.2302.07842} {Augmented language models: a survey}.
\newblock \emph{CoRR}, abs/2302.07842.

\bibitem[{Nie et~al.(2023)Nie, Zhang, Wang, and Liu}]{kb-coder}
Zhijie Nie, Richong Zhang, Zhongyuan Wang, and Xudong Liu. 2023.
\newblock \href {https://doi.org/10.48550/ARXIV.2309.04695} {Code-style in-context learning for knowledge-based question answering}.
\newblock \emph{CoRR}, abs/2309.04695.

\bibitem[{OpenAI(2023{\natexlab{a}})}]{gpt4}
OpenAI. 2023{\natexlab{a}}.
\newblock \href {https://doi.org/10.48550/arXiv.2303.08774} {{GPT-4} technical report}.
\newblock \emph{CoRR}, abs/2303.08774.

\bibitem[{OpenAI(2023{\natexlab{b}})}]{gpt35turbo}
OpenAI. 2023{\natexlab{b}}.
\newblock Models - {OpenAI API}.
\newblock \url{https://platform.openai.com/docs/models/gpt-3-5}.

\bibitem[{Pourreza and Rafiei(2023)}]{din-sql}
Mohammadreza Pourreza and Davood Rafiei. 2023.
\newblock \href {http://papers.nips.cc/paper\_files/paper/2023/hash/72223cc66f63ca1aa59edaec1b3670e6-Abstract-Conference.html} {{DIN-SQL:} decomposed in-context learning of text-to-sql with self-correction}.
\newblock In \emph{Advances in Neural Information Processing Systems 36: Annual Conference on Neural Information Processing Systems 2023, NeurIPS 2023, New Orleans, LA, USA, December 10 - 16, 2023}.

\bibitem[{Qin et~al.(2023{\natexlab{a}})Qin, Hu, Lin, Chen, Ding, Cui, Zeng, Huang, Xiao, Han, Fung, Su, Wang, Qian, Tian, Zhu, Liang, Shen, Xu, Zhang, Ye, Li, Tang, Yi, Zhu, Dai, Yan, Cong, Lu, Zhao, Huang, Yan, Han, Sun, Li, Phang, Yang, Wu, Ji, Liu, and Sun}]{toollearning}
Yujia Qin, Shengding Hu, Yankai Lin, Weize Chen, Ning Ding, Ganqu Cui, Zheni Zeng, Yufei Huang, Chaojun Xiao, Chi Han, Yi~Ren Fung, Yusheng Su, Huadong Wang, Cheng Qian, Runchu Tian, Kunlun Zhu, Shihao Liang, Xingyu Shen, Bokai Xu, Zhen Zhang, Yining Ye, Bowen Li, Ziwei Tang, Jing Yi, Yuzhang Zhu, Zhenning Dai, Lan Yan, Xin Cong, Yaxi Lu, Weilin Zhao, Yuxiang Huang, Junxi Yan, Xu~Han, Xian Sun, Dahai Li, Jason Phang, Cheng Yang, Tongshuang Wu, Heng Ji, Zhiyuan Liu, and Maosong Sun. 2023{\natexlab{a}}.
\newblock \href {https://doi.org/10.48550/arXiv.2304.08354} {Tool learning with foundation models}.
\newblock \emph{CoRR}, abs/2304.08354.

\bibitem[{Qin et~al.(2023{\natexlab{b}})Qin, Liang, Ye, Zhu, Yan, Lu, Lin, Cong, Tang, Qian, Zhao, Tian, Xie, Zhou, Gerstein, Li, Liu, and Sun}]{toolllm}
Yujia Qin, Shihao Liang, Yining Ye, Kunlun Zhu, Lan Yan, Yaxi Lu, Yankai Lin, Xin Cong, Xiangru Tang, Bill Qian, Sihan Zhao, Runchu Tian, Ruobing Xie, Jie Zhou, Mark Gerstein, Dahai Li, Zhiyuan Liu, and Maosong Sun. 2023{\natexlab{b}}.
\newblock \href {https://doi.org/10.48550/ARXIV.2307.16789} {{ToolLLM}: Facilitating large language models to master 16000+ real-world apis}.
\newblock \emph{CoRR}, abs/2307.16789.

\bibitem[{Rajkumar et~al.(2022)Rajkumar, Li, and Bahdanau}]{rajkumar2022api}
Nitarshan Rajkumar, Raymond Li, and Dzmitry Bahdanau. 2022.
\newblock \href {https://doi.org/10.48550/ARXIV.2204.00498} {Evaluating the text-to-sql capabilities of large language models}.
\newblock \emph{CoRR}, abs/2204.00498.

\bibitem[{Reimers and Gurevych(2019)}]{reimers-gurevych-2019-sentence}
Nils Reimers and Iryna Gurevych. 2019.
\newblock \href {https://doi.org/10.18653/v1/D19-1410} {Sentence-{BERT}: Sentence embeddings using {S}iamese {BERT}-networks}.
\newblock In \emph{Proceedings of the 2019 Conference on Empirical Methods in Natural Language Processing and the 9th International Joint Conference on Natural Language Processing (EMNLP-IJCNLP)}, pages 3982--3992, Hong Kong, China. Association for Computational Linguistics.

\bibitem[{Schick et~al.(2023)Schick, Dwivedi{-}Yu, Dess{\`{\i}}, Raileanu, Lomeli, Zettlemoyer, Cancedda, and Scialom}]{toolformer}
Timo Schick, Jane Dwivedi{-}Yu, Roberto Dess{\`{\i}}, Roberta Raileanu, Maria Lomeli, Luke Zettlemoyer, Nicola Cancedda, and Thomas Scialom. 2023.
\newblock \href {https://doi.org/10.48550/arXiv.2302.04761} {Toolformer: Language models can teach themselves to use tools}.
\newblock \emph{CoRR}, abs/2302.04761.

\bibitem[{Shridhar et~al.(2021)Shridhar, Yuan, C{\^{o}}t{\'{e}}, Bisk, Trischler, and Hausknecht}]{alfworld}
Mohit Shridhar, Xingdi Yuan, Marc{-}Alexandre C{\^{o}}t{\'{e}}, Yonatan Bisk, Adam Trischler, and Matthew~J. Hausknecht. 2021.
\newblock \href {https://openreview.net/forum?id=0IOX0YcCdTn} {Alfworld: Aligning text and embodied environments for interactive learning}.
\newblock In \emph{9th International Conference on Learning Representations, {ICLR} 2021, Virtual Event, Austria, May 3-7, 2021}. OpenReview.net.

\bibitem[{Song et~al.(2023)Song, Wu, Washington, Sadler, Chao, and Su}]{song2023llmplanner}
Chan~Hee Song, Jiaman Wu, Clayton Washington, Brian~M. Sadler, Wei-Lun Chao, and Yu~Su. 2023.
\newblock {LLM-Planner}: Few-shot grounded planning for embodied agents with large language models.
\newblock In \emph{Proceedings of the IEEE/CVF International Conference on Computer Vision (ICCV)}.

\bibitem[{Su(2023)}]{su2023language}
Yu~Su. 2023.
\newblock \href {https://yusu.substack.com/p/language-agents} {Language agents: a critical evolutionary step of artificial intelligence}.
\newblock \emph{yusu.substack.com}.

\bibitem[{Su et~al.(2016)Su, Sun, Sadler, Srivatsa, Gur, Yan, and Yan}]{graphq}
Yu~Su, Huan Sun, Brian~M. Sadler, Mudhakar Srivatsa, Izzeddin Gur, Zenghui Yan, and Xifeng Yan. 2016.
\newblock \href {https://doi.org/10.18653/V1/D16-1054} {On generating characteristic-rich question sets for {QA} evaluation}.
\newblock In \emph{Proceedings of the 2016 Conference on Empirical Methods in Natural Language Processing, {EMNLP} 2016, Austin, Texas, USA, November 1-4, 2016}, pages 562--572. The Association for Computational Linguistics.

\bibitem[{Sun et~al.(2023)Sun, Zhang, Yan, Gao, Ong, Chen, and Su}]{sun-etal-2023-battle}
Shuo Sun, Yuchen Zhang, Jiahuan Yan, Yuze Gao, Donovan Ong, Bin Chen, and Jian Su. 2023.
\newblock \href {https://doi.org/10.18653/v1/2023.findings-emnlp.750} {Battle of the large language models: Dolly vs {LL}a{MA} vs vicuna vs guanaco vs bard vs {C}hat{GPT} - a text-to-{SQL} parsing comparison}.
\newblock In \emph{Findings of the Association for Computational Linguistics: EMNLP 2023}, pages 11225--11238, Singapore. Association for Computational Linguistics.

\bibitem[{Tai et~al.(2023)Tai, Chen, Zhang, Deng, and Sun}]{tai2023exploring}
Chang{-}Yu Tai, Ziru Chen, Tianshu Zhang, Xiang Deng, and Huan Sun. 2023.
\newblock \href {https://aclanthology.org/2023.emnlp-main.327} {Exploring chain of thought style prompting for text-to-sql}.
\newblock In \emph{Proceedings of the 2023 Conference on Empirical Methods in Natural Language Processing, {EMNLP} 2023, Singapore, December 6-10, 2023}, pages 5376--5393. Association for Computational Linguistics.

\bibitem[{Talmor and Berant(2018)}]{cwq}
Alon Talmor and Jonathan Berant. 2018.
\newblock \href {https://doi.org/10.18653/V1/N18-1059} {The web as a knowledge-base for answering complex questions}.
\newblock In \emph{Proceedings of the 2018 Conference of the North American Chapter of the Association for Computational Linguistics: Human Language Technologies, {NAACL-HLT} 2018, New Orleans, Louisiana, USA, June 1-6, 2018, Volume 1 (Long Papers)}, pages 641--651. Association for Computational Linguistics.

\bibitem[{Touvron et~al.(2023)Touvron, Martin, Stone, Albert, Almahairi, Babaei, Bashlykov, Batra, Bhargava, Bhosale, Bikel, Blecher, Canton{-}Ferrer, Chen, Cucurull, Esiobu, Fernandes, Fu, Fu, Fuller, Gao, Goswami, Goyal, Hartshorn, Hosseini, Hou, Inan, Kardas, Kerkez, Khabsa, Kloumann, Korenev, Koura, Lachaux, Lavril, Lee, Liskovich, Lu, Mao, Martinet, Mihaylov, Mishra, Molybog, Nie, Poulton, Reizenstein, Rungta, Saladi, Schelten, Silva, Smith, Subramanian, Tan, Tang, Taylor, Williams, Kuan, Xu, Yan, Zarov, Zhang, Fan, Kambadur, Narang, Rodriguez, Stojnic, Edunov, and Scialom}]{llama2}
Hugo Touvron, Louis Martin, Kevin Stone, Peter Albert, Amjad Almahairi, Yasmine Babaei, Nikolay Bashlykov, Soumya Batra, Prajjwal Bhargava, Shruti Bhosale, Dan Bikel, Lukas Blecher, Cristian Canton{-}Ferrer, Moya Chen, Guillem Cucurull, David Esiobu, Jude Fernandes, Jeremy Fu, Wenyin Fu, Brian Fuller, Cynthia Gao, Vedanuj Goswami, Naman Goyal, Anthony Hartshorn, Saghar Hosseini, Rui Hou, Hakan Inan, Marcin Kardas, Viktor Kerkez, Madian Khabsa, Isabel Kloumann, Artem Korenev, Punit~Singh Koura, Marie{-}Anne Lachaux, Thibaut Lavril, Jenya Lee, Diana Liskovich, Yinghai Lu, Yuning Mao, Xavier Martinet, Todor Mihaylov, Pushkar Mishra, Igor Molybog, Yixin Nie, Andrew Poulton, Jeremy Reizenstein, Rashi Rungta, Kalyan Saladi, Alan Schelten, Ruan Silva, Eric~Michael Smith, Ranjan Subramanian, Xiaoqing~Ellen Tan, Binh Tang, Ross Taylor, Adina Williams, Jian~Xiang Kuan, Puxin Xu, Zheng Yan, Iliyan Zarov, Yuchen Zhang, Angela Fan, Melanie Kambadur, Sharan Narang, Aur{\'{e}}lien Rodriguez, Robert Stojnic, Sergey Edunov,
  and Thomas Scialom. 2023.
\newblock \href {https://doi.org/10.48550/arXiv.2307.09288} {Llama 2: Open foundation and fine-tuned chat models}.
\newblock \emph{CoRR}, abs/2307.09288.

\bibitem[{Wang et~al.(2024)Wang, Fried, and Neubig}]{trove}
Zhiruo Wang, Daniel Fried, and Graham Neubig. 2024.
\newblock \href {https://doi.org/10.48550/ARXIV.2401.12869} {Trove: Inducing verifiable and efficient toolboxes for solving programmatic tasks}.
\newblock \emph{CoRR}, abs/2401.12869.

\bibitem[{Wei et~al.(2022)Wei, Wang, Schuurmans, Bosma, Ichter, Xia, Chi, Le, and Zhou}]{CoT}
Jason Wei, Xuezhi Wang, Dale Schuurmans, Maarten Bosma, Brian Ichter, Fei Xia, Ed~H. Chi, Quoc~V. Le, and Denny Zhou. 2022.
\newblock \href {http://papers.nips.cc/paper\_files/paper/2022/hash/9d5609613524ecf4f15af0f7b31abca4-Abstract-Conference.html} {Chain-of-thought prompting elicits reasoning in large language models}.
\newblock In \emph{NeurIPS}.

\bibitem[{Yao et~al.(2022)Yao, Zhao, Yu, Du, Shafran, Narasimhan, and Cao}]{react}
Shunyu Yao, Jeffrey Zhao, Dian Yu, Nan Du, Izhak Shafran, Karthik Narasimhan, and Yuan Cao. 2022.
\newblock \href {https://doi.org/10.48550/arXiv.2210.03629} {{ReAct}: Synergizing reasoning and acting in language models}.
\newblock \emph{CoRR}, abs/2210.03629.

\bibitem[{Yih et~al.(2016)Yih, Richardson, Meek, Chang, and Suh}]{yih-etal-2016-value}
Wen-tau Yih, Matthew Richardson, Chris Meek, Ming-Wei Chang, and Jina Suh. 2016.
\newblock \href {https://doi.org/10.18653/v1/P16-2033} {The value of semantic parse labeling for knowledge base question answering}.
\newblock In \emph{Proceedings of the 54th Annual Meeting of the Association for Computational Linguistics (Volume 2: Short Papers)}, pages 201--206, Berlin, Germany. Association for Computational Linguistics.

\bibitem[{Yu et~al.(2023)Yu, Zhang, Ng, Zhu, Li, Wang, Hu, Wang, Wang, and Xiang}]{decaf}
Donghan Yu, Sheng Zhang, Patrick Ng, Henghui Zhu, Alexander~Hanbo Li, Jun Wang, Yiqun Hu, William~Yang Wang, Zhiguo Wang, and Bing Xiang. 2023.
\newblock \href {https://openreview.net/pdf?id=XHc5zRPxqV9} {{DecAF}: Joint decoding of answers and logical forms for question answering over knowledge bases}.
\newblock In \emph{The Eleventh International Conference on Learning Representations, {ICLR} 2023, Kigali, Rwanda, May 1-5, 2023}. OpenReview.net.

\bibitem[{Yu et~al.(2018)Yu, Zhang, Yang, Yasunaga, Wang, Li, Ma, Li, Yao, Roman, Zhang, and Radev}]{spider}
Tao Yu, Rui Zhang, Kai Yang, Michihiro Yasunaga, Dongxu Wang, Zifan Li, James Ma, Irene Li, Qingning Yao, Shanelle Roman, Zilin Zhang, and Dragomir~R. Radev. 2018.
\newblock \href {https://doi.org/10.18653/v1/d18-1425} {{Spider}: {A} large-scale human-labeled dataset for complex and cross-domain semantic parsing and text-to-sql task}.
\newblock In \emph{Proceedings of the 2018 Conference on Empirical Methods in Natural Language Processing, Brussels, Belgium, October 31 - November 4, 2018}, pages 3911--3921. Association for Computational Linguistics.

\bibitem[{Zhang et~al.(2018)Zhang, Dai, Kozareva, Smola, and Song}]{metaqa}
Yuyu Zhang, Hanjun Dai, Zornitsa Kozareva, Alexander~J. Smola, and Le~Song. 2018.
\newblock \href {https://doi.org/10.1609/AAAI.V32I1.12057} {Variational reasoning for question answering with knowledge graph}.
\newblock In \emph{Proceedings of the Thirty-Second {AAAI} Conference on Artificial Intelligence, (AAAI-18), the 30th innovative Applications of Artificial Intelligence (IAAI-18), and the 8th {AAAI} Symposium on Educational Advances in Artificial Intelligence (EAAI-18), New Orleans, Louisiana, USA, February 2-7, 2018}, pages 6069--6076. {AAAI} Press.

\bibitem[{Zheng et~al.(2023)Zheng, Chiang, Sheng, Zhuang, Wu, Zhuang, Lin, Li, Li, Xing, Zhang, Gonzalez, and Stoica}]{chatbot_arena}
Lianmin Zheng, Wei-Lin Chiang, Ying Sheng, Siyuan Zhuang, Zhanghao Wu, Yonghao Zhuang, Zi~Lin, Zhuohan Li, Dacheng Li, Eric.~P Xing, Hao Zhang, Joseph~E. Gonzalez, and Ion Stoica. 2023.
\newblock \href {http://arxiv.org/abs/2306.05685} {Judging {LLM-as-a-judge} with {MT-Bench} and {Chatbot Arena}}.

\bibitem[{Zhong et~al.(2017)Zhong, Xiong, and Socher}]{wikisql}
Victor Zhong, Caiming Xiong, and Richard Socher. 2017.
\newblock \href {http://arxiv.org/abs/1709.00103} {{Seq2SQL}: Generating structured queries from natural language using reinforcement learning}.
\newblock \emph{CoRR}, abs/1709.00103.

\end{thebibliography}
\bibliographystyle{acl_natbib}
\clearpage
\appendix


\setcounter{table}{0}
\renewcommand\thetable{\Alph{section}.\arabic{table}}
\setcounter{figure}{0}
\renewcommand\thefigure{\Alph{section}.\arabic{figure}}

\section*{Appendices}
In this supplementary material, we provide further details as follows:
\begin{itemize}[topsep=2pt,itemsep=2pt,partopsep=2pt, parsep=2pt]
\item \autoref{subsec:detailed tool definitions}: Detailed Tool Definitions
\item \autoref{appendix:benchmarks}: Benchmark Statistics
\item \autoref{subsec:detailed prompts}: Prompts
\item \autoref{appendix:exp}: Additional Results
\end{itemize}

\section{Detailed Tool Definitions}
\label{subsec:detailed tool definitions}
In this section, we detail the descriptions of our customized tools for both environments.
Specifically, we implement \num{12} different tools for databases and \num{7} different tools for KBs.
The tool selection is carefully made based on our domain knowledge of these environments.
Note that, for databases, we direct prompt the LLM with the DB schema information in API docs format~\cite{rajkumar2022api}, as a result, our tools focus on helping the LLM better engage with the database content.

\tcbset{
    enhanced,
    colback=white, 
    colframe=black!50!black, 
    fonttitle=\small\bfseries, 
    colbacktitle=black!60!black,
    coltitle=white, 
    attach boxed title to top left={yshift=-2mm, xshift=2mm},
    left=1mm, 
    right=1mm,
    top=2mm, 
    bottom=1mm,
    breakable, 
    boxsep=2pt, 
    fontupper=\small, 
    fontlower=\small,
    unbreakable
}

\subsection{Databases}
\label{appendix:db_tools}
Navigational tools for databases:

\begin{tcolorbox}[title=\texttt{find\_columns\_containing\_value(value)}]
This function can help to find columns that contain the given cell value, which can help you make better decisions in decoding the right column to use. Note that, the value here means cell value in the rows of the column, not the column name. \\
\textbf{Prerequisite:} n/a
\end{tcolorbox}

\begin{tcolorbox}[title=\texttt{find\_columns\_containing\_value\_fuzzy(value)}]
Sometimes find\_columns\_containing\_cell\_value may not find a column with the exact matched cell value. This function can help to find columns that potentially contain the target cell value with fuzzy matching. Note that, the value here means cell value in the rows of the column, not the column name. \\
\textbf{Prerequisite:} n/a
\end{tcolorbox}

\begin{tcolorbox}[title=\texttt{get\_distinct\_values(table, column)}]
Returns the distinct values in the given column. This may mainly help you make better decisions in decoding the right value to use. \\
\textbf{Prerequisite:} n/a
\end{tcolorbox}

\begin{tcolorbox}[title=\texttt{is\_value\_in\_column(table, column, value)}]
Returns whether the given value is in the given column. You can use this function to better detect the right column to use. \\
\textbf{Prerequisite:} n/a
\end{tcolorbox}

\begin{tcolorbox}[title=\texttt{get\_date\_format(table, column)}]
Returns an example item of the given Date column. This may help you to better understand the date format in the column. \\
\textbf{Prerequisite:} n/a
\end{tcolorbox}

\begin{tcolorbox}[title=\texttt{search\_by\_SQL(query)}]
Executing a SQL query to search the table. \\
\textbf{Prerequisite:} n/a
\end{tcolorbox}

Functional tools for databases:

\begin{tcolorbox}[title=\texttt{from(from\_statement)}]
This function specifies the FROM clause, e.g., from("FROM table1") or from("FROM table1 JOIN table2 ON table1.id = table2.id") \\
\textbf{Prerequisite:} n/a
\end{tcolorbox}

\begin{tcolorbox}[title=\texttt{where(where\_statement)}]
This function specifies the WHERE clause, e.g., where("WHERE table1.id = 1"). \\
\textbf{Prerequisite:} from
\end{tcolorbox}

\begin{tcolorbox}[title=\texttt{select(select\_statement)}]
This function specifies the SELECT clause, e.g., select("SELECT table1.id"). \\
\textbf{Prerequisite:} from, where
\end{tcolorbox}

\begin{tcolorbox}[title=\texttt{group\_by(group\_by\_statement)}]
This function specifies the GROUP BY clause, e.g., group\_by("GROUP BY table1.id"). \\
\textbf{Prerequisite:} from, where, select
\end{tcolorbox}

\begin{tcolorbox}[title=\texttt{having(having\_statement)}]
This function specifies the HAVING clause, e.g., having("HAVING table1.id = 1"). \\
\textbf{Prerequisite:} from, where, select, group\_by
\end{tcolorbox}

\begin{tcolorbox}[title=\texttt{order\_by(order\_by\_statement)}]
This function specifies an additional constraint like ordering. For example, order\_by("ORDER BY table1.id DESC LIMIT 3"). \\
\textbf{Prerequisite:} from, where, select
\end{tcolorbox}

\subsection{Knowledge Bases}
\label{appendix:kb_tools}
Navigational tools for KBs:

\begin{tcolorbox}[title=\texttt{get\_relations(variable) -> list of relations}]
A variable can be either an entity or a set of entities (i.e., the result of a previous query). This function helps to navigate all relations in the KB connected to the variable, so you can decide which relation is the most useful to find the answer to the question. \\
A simple use case can be `get\_relations(Barack Obama)', which finds all relations/edges starting from the entity Barack Obama. \\
The argument of get\_relations should always be an entity or a variable (e.g., \#0) and not anything else. \\
\textbf{Prerequisite:} n/a
\end{tcolorbox}

\begin{tcolorbox}[title=\texttt{get\_neighbors(v, r) -> variable}]
Given a variable, this function returns all entities connected to the variable via the given relation. Note that, get\_neighbors() can only be used after get\_relations() is used to find a set of viable relations. \\
A simple use case can be `get\_neighbors(Barack Obama, people.person.profession)', which returns the profession of Obama in Freebase. \\
\textbf{Prerequisite:} get\_relations
\end{tcolorbox}

\begin{tcolorbox}[title=\texttt{get\_attributes(v) -> list of attributes}]
This function helps to find all numerical attributes of the variable. Please only use it if the question seeks for a superlative accumulation (i.e., argmax or argmin). \\
\textbf{Prerequisite:} get\_neighbors
\end{tcolorbox}

Functional tools for KBs:

\begin{tcolorbox}[title=\texttt{argmax(v, a) -> variable}]
Given a variable, this function returns the entity with the maximum value of the given attribute. It can only be used after get\_attributes() is used to find a set of viable attributes. \\
A simple use case can be `argmax(variable, age)', which returns the oldest entity belonging to the variable. \\
\textbf{Prerequisite:} get\_attributes
\end{tcolorbox}

\begin{tcolorbox}[title=\texttt{argmin(v, a) -> variable}]
Given a variable, this function returns the entity with the minimum value of the given attribute. It can only be used after get\_attributes() is used to find a set of viable attributes. \\
A simple use case can be `argmin(variable, age)', which returns the youngest entity belonging to the variable. \\
\textbf{Prerequisite:} get\_attributes
\end{tcolorbox}

\begin{tcolorbox}[title=\texttt{intersection(v1, v2) -> variable}]
Given two variables, this function returns the intersection of the two variables. The two variables must be of the same type. \\
\textbf{Prerequisite:} get\_neighbors
\end{tcolorbox}


\begin{tcolorbox}[title=\texttt{count(v) -> int}]
Given a variable, this function returns the number of entities belonging to the variable. \\
\textbf{Prerequisite:} get\_neighbors
\end{tcolorbox}

\section{Benchmark Statistics}
\label{appendix:benchmarks}
\begin{table*}[!htb]
\small
\begin{subtable}{\linewidth}
    \small
    \centering
    \resizebox{0.8\textwidth}{!}{
    \begin{tabular}{lcccc}
    \toprule
    \textbf{Dataset} & \textbf{\# Table/DB} & \textbf{\# Row/DB} & \textbf{\% Require Cont.} \\
    \midrule
    \WikiSQL~\cite{wikisql} & \num{1} & \num{17} & \num{0.0}\\
    \Spider~\cite{spider} & \num{5.1} & \num{2}K & \num{0.0} \\
    \Bird~\cite{bird} & \num{7.3} & \num{549}K& \num{32.3} \\
    \bottomrule
    \end{tabular}}
    \caption{Databases}
    \label{table:bird}
\end{subtable}

\begin{subtable}{\linewidth}
    \small
    \centering
    \resizebox{0.9\textwidth}{!}{
    \begin{tabular}{lccccc}
    \toprule
    \textbf{Dataset} & \textbf{\# Relations/KB} & \textbf{\# Triples/KB} & \textbf{\# Hops} & \textbf{\% Have Aggr.} \\
    \midrule
    \MetaQ~\cite{metaqa} & \num{9} & \num{135}K & \num{2.1} & \num{0.0}\\
    \WebQSP~\cite{yih-etal-2016-value} & \num{19}K& \num{3}B & \num{1.5} & \num{4.9} \\
    \GrailQ~\cite{grailqa}& \num{19}K& \num{3}B& \num{1.4} & \num{18.5} \\
    \KBQAAgent (Ours) &\num{19}K & \num{3}B& \num{2.9} & \num{38.4}\\
    \bottomrule
    \end{tabular}
    }
    \caption{Knowledge Bases}
    \label{table:kb-agent}
\end{subtable}
\caption{Our curated benchmarks more accurately mirror real-world complexity, offering a more effective assessment of language agents. Aggr. denotes aggregation functions.}
\label{table:benchmark statistics}
\end{table*}
In Table \ref{table:benchmark statistics}, we present the statistics of \Bird and \KBQAAgent, which we have chosen for our evaluation. Relative to established benchmarks in text-to-SQL parsing and KBQA, \Bird and \KBQAAgent exhibit significantly greater complexity, making them more suitable for assessing the capabilities of language agents.

\section{Prompts}
\label{subsec:detailed prompts}
\begin{figure*}[!ht]
    \centering
    \includegraphics[width=\textwidth]{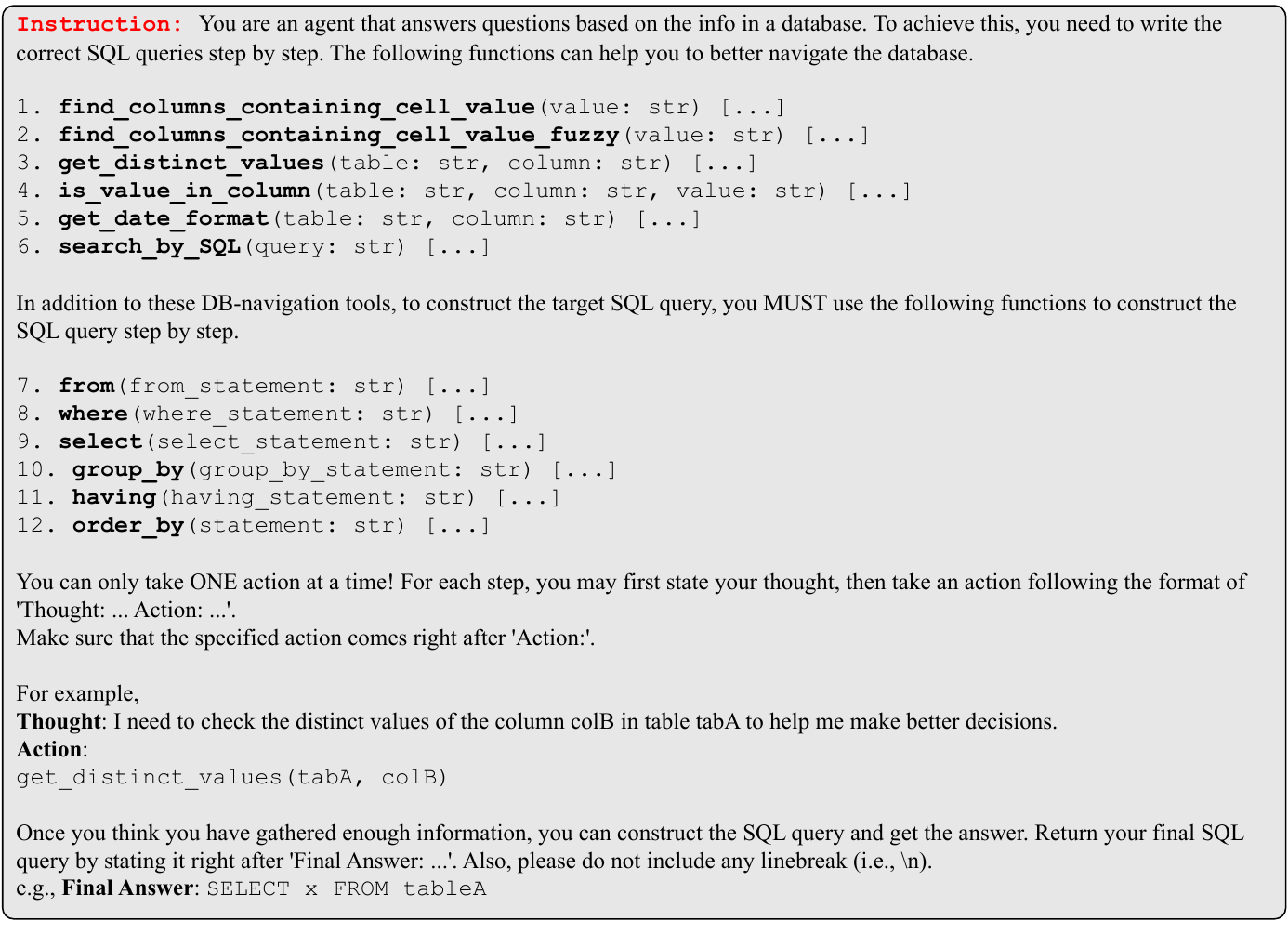}
    \caption{Instructions for using database tools. Descriptions of tools are omitted.}
\label{figure:db tools}
\end{figure*}
\begin{figure*}[h]
\centering
\includegraphics[width=\textwidth]{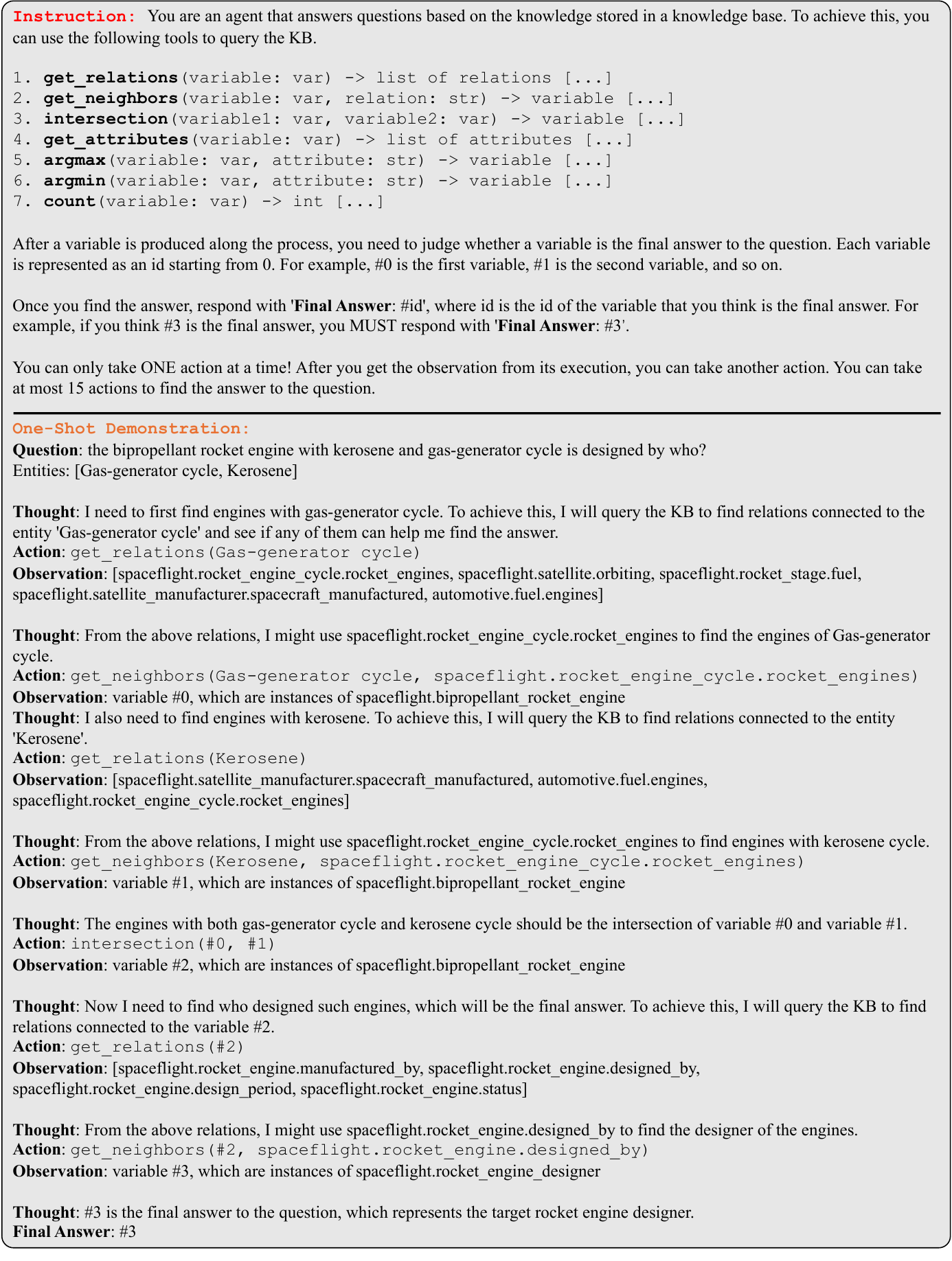}
\caption{Instructions and a one-shot demonstration for using KB tools. Descriptions of tools are omitted.}
\label{figure:kb tools}
\end{figure*}
\begin{figure*}[h]
\centering
\includegraphics[width=\textwidth]{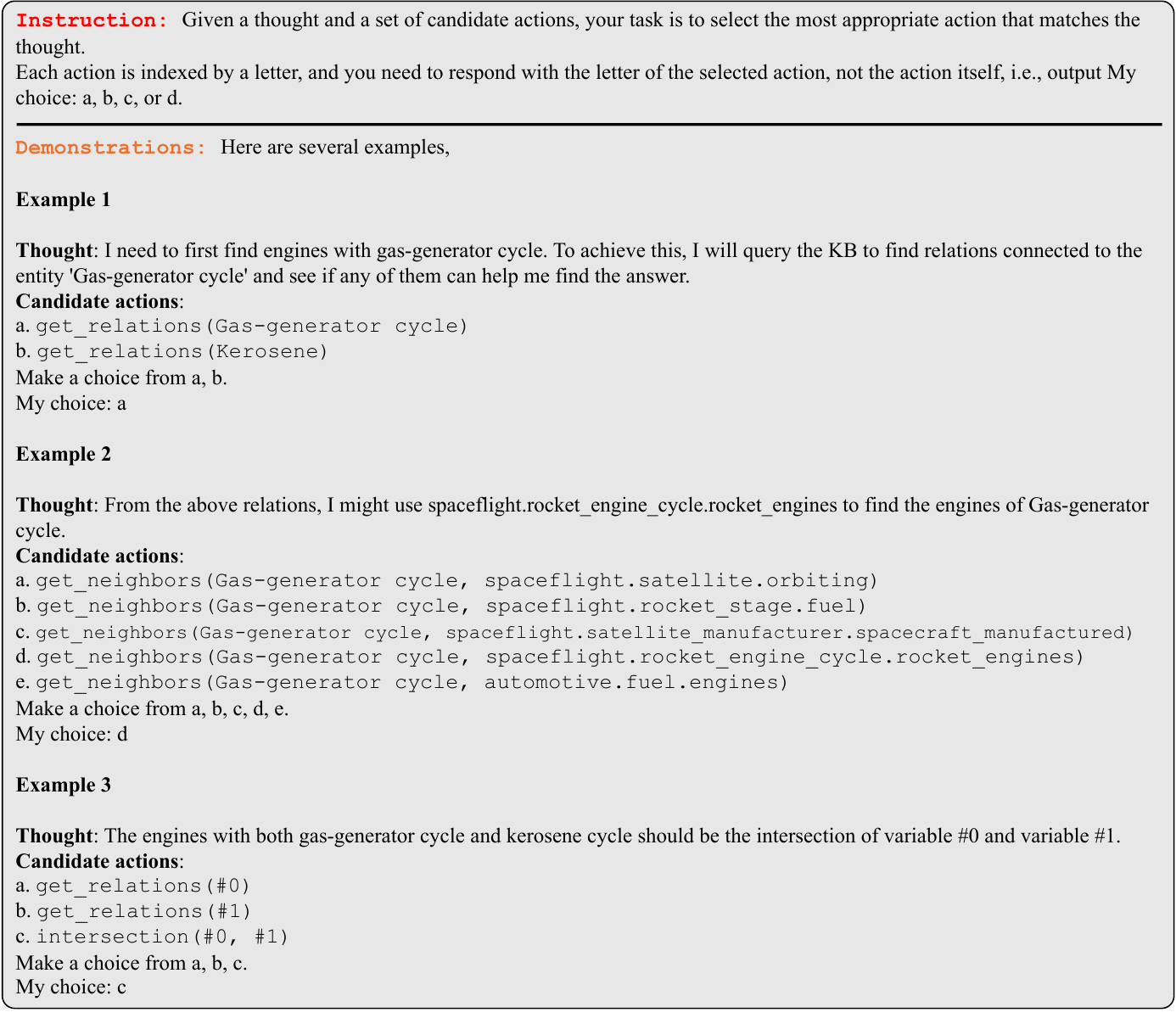}
\caption{Prompt for candidate action selection in decoupled generation for KB.}
\label{figure:kb decoupled generation}
\end{figure*}

Instructions and demonstrations for using database tools are shown in Figure \ref{figure:db tools}. 
Note that, we also include the schema information of the database in API Docs in our prompt, which is not shown here.
This design choice has been a common practice for text-to-SQL parsing with LLMs~\cite{tai2023exploring, sun-etal-2023-battle}.
Instructions and demonstrations for using KB tools are shown in Figure \ref{figure:kb tools}. 
The instruction and demonstration for candidate selection in \textit{decoupled generation} for KB is shown in Figure \ref{figure:kb decoupled generation}.
Additionally, we also show an example of input we use for our KB experiments in Section~\ref{sec:exp_middleware}.
For the input used for databases in Section~\ref{sec:exp_middleware}, we strictly follow the standard way of prompting with API docs plus exemplar rows~\cite{bird, rajkumar2022api}.



\lstset{
    basicstyle=\small\ttfamily, 
    breaklines=true,  
    frame=single,        
    framesep=10pt,      
    breakindent=0pt,
    xleftmargin=10pt,  
    xrightmargin=10pt,
    escapeinside={(*@}{@*)}
}

\begin{figure*}[h]
\centering
\includegraphics[width=\textwidth]{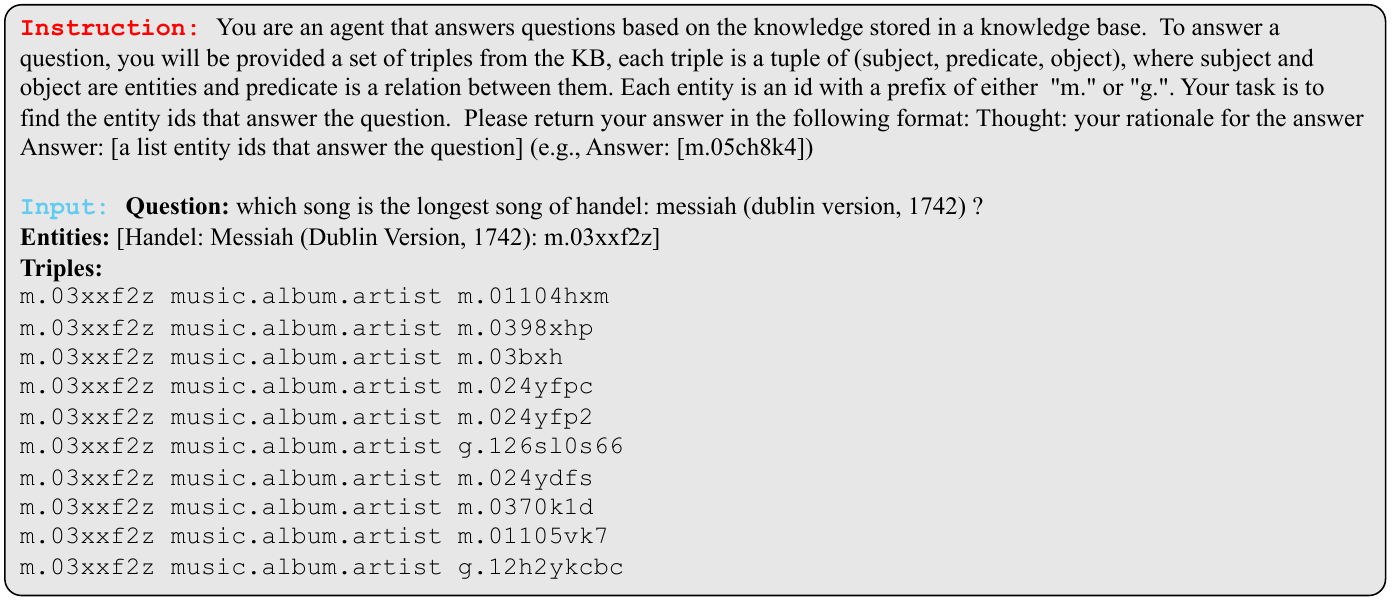}
\caption{Input for question ``\textit{which song is the longest song of handel: messiah (dublin version, 1742)?}" with \num{10} triples sampled from the KB, which is used in Section~\ref{sec:exp_middleware}.}
\label{figure:kg_triples}
\end{figure*}

\section{Additional Results}
\label{appendix:exp}
\begin{table*}
    \small
    \centering
    \begin{subtable}{\linewidth}
        \small
        \centering
        \resizebox{0.9\textwidth}{!}{
        \begin{tabular}{lccccc}
            \toprule
            & Avg Input Tokens & Avg Time (s) & Avg Rounds & Perceivable Rows & EX (100 questions) \\
            \midrule
            \OurMethod (\textit{error feedback}) & \num{14257.2} & \num{15.5} & \num{7.1} & \(\infty\) & \num{39.0} \\
            API Docs Prompt & \num{1296.8} & \num{2.5} & \num{1.0} & \num{0} & \num{12.0} \\
            API Docs Prompt (10 rows) & \num{4635.5} & \num{2.9} & \num{1.0} & \num{10} & \num{22.0} \\
            API Docs Prompt (20 rows) & \num{8459.4} & \num{2.9} & \num{1.0} & \num{20} & \num{9.0} \\
            \bottomrule
        \end{tabular}}
        \caption{Table 1}
    \end{subtable}
    
    
    \begin{subtable}{\textwidth}
        \centering
        \small
        \resizebox{0.8\textwidth}{!}{
        \begin{tabular}{lccccc}
            \toprule
            & Avg Input Tokens & Avg Time (s) & Avg Rounds & Perceivable Triples & F1 \\
            \midrule
            \OurMethod (\textit{error feedback}) & \num{25140.7} & \num{22.1} & \num{9.8} & \(\infty\) & \num{55.1} \\
            \OurMethod (\textit{decoupled generation}) & \num{23331.8} & \num{20.9} & \num{9.3} & \(\infty\) & \num{59.3} \\
            Pangu-Chat & \num{4675.7} & \num{67.3} & \num{4.5} & \(\infty\) & \num{27.1} \\
            Direct Prompt (200 triples) & \num{7654.5} & \num{3.1} & \num{1.0} & \num{200} & \num{0.8} \\
            \bottomrule
        \end{tabular}}
        \caption{Table 2}
    \end{subtable}

    \caption{The average running time and input tokens for different methods. Note that, no existing research addresses the challenge of handling extensive DB content, limiting existing methods to perceiving only a small number of rows. This is a critical gap in the literature that we aim to fill.}
    \label{tab:efficiency}
\end{table*}

\begin{table*}
    \centering
    \resizebox{0.7\textwidth}{!}{
    \begin{tabular}{lcc}
        \toprule
        Setting & gpt-3.5-turbo-0613 & gpt-4-0613 \\
        \midrule
        \textit{w/} all & \num{14.0} & \num{38.0} \\
        \textit{w/o} \texttt{find\_columns\_containing\_value} & \num{14.0} & \num{38.0} \\
        \textit{w/o} \texttt{find\_columns\_containing\_value\_fuzzy} & \num{12.0} & \num{34.0} \\
        \textit{w/o} \texttt{get\_distinct\_values} & \num{10.0} & \num{32.0} \\
        \textit{w/o} \texttt{is\_value\_in\_column} & \num{14.0} & \num{36.0} \\
        \textit{w/o} all four & \num{8.0} & \num{22.0} \\
        \bottomrule
    \end{tabular}
    }
    \caption{Ablation study on different tools used in our DB tasks.}
    \label{tab:ablation}
\end{table*}
\subsection{Comparing with DIN-SQL and DAIL-SQL}
\label{appendix:exp_comp}
In addition to API Docs Prompt, we also compare with two strong baselines with their source code available: DIN-SQL~\cite{din-sql} and DAIL-SQL~\cite{dial-sql} from \Bird's leaderboard.
Note that in their original submission, they only evaluated under the oracle knowledge setting. 
To ensure a fair comparison with \OurMethod under the same without oracle knowledge setting, we adapted their source code to evaluate without oracle knowledge with minimal changes. 
Specifically, some modules in their pipeline design require several in-context demonstrations, which we also preserved.
Consequently, we use 6-shot for DIN-SQL and 7-shot for DAIL-SQL, following their original codebase.
In particular, we randomly sample \num{100} questions from \Bird's dev set and evaluate them on these questions.
The results is presented in Table~\ref{tab:comparison}.

\subsection{Efficiency Analysis}
We further look into the efficiency (\eg, \textit{avg.} tokens and \textit{avg.} running time) of different methods to gain a deeper understanding.
For DBs, we compare \OurMethod with the most straightforward single-round baseline API Docs Prompt.
For KBs, we compare \OurMethod with Pangu, as all other baselines significantly underperform.
The concrete results are presented in Table~\ref{tab:efficiency}.

\subsection{Ablation for Tools}
For KBQA, an ablation experiment for tools would be trivial as each tool deterministically contributes to the final KB queries. For example, removing \texttt{get\_relations} or \texttt{get\_neighbors} would yield 0\% performance on all questions, while removing \texttt{count} would result in 0\% performance on counting questions.
As a result, we only do the ablation experiment for text-to-SQL parsing. 
Specifically, we first identify the top-4 commonly used tools based on GPT-4's predictions: \texttt{find\_columns\_containing\_value}, \texttt{find\_columns\_containing\_value\_fuzzy}, \texttt{get\_distinct\_values}, and \texttt{is\_value\_in\_column}. 
On 50 out of the previous 100 questions (due to the budget concern), we show the results in Table~\ref{tab:ablation}.
An intriguing observation is the robustness exhibited when removing only one tool, owing to the redundancy built into our tool designs. Nevertheless, when all four tools were removed, the performance experienced a substantial decline.

\begin{table}[!tb]
    \centering
    \resizebox{0.5\textwidth}{!}{
    \begin{tabular}{lcc}
        \toprule
        & gpt-3.5-turbo-0613 & gpt-4-0613 \\
        \midrule
        \OurMethod (0-shot) & \textbf{\num{16.0}} & \textbf{\num{39.0}} \\
        API Doc Prompt (0-shot) & \num{6.0} & \num{12.0} \\
        DIN-SQL (6-shot) & - & \num{22.0} \\
        DAIL-SQL (7-shot) & \num{8.0} & \num{11.0} \\
        \bottomrule
    \end{tabular}
    }
    \caption{Results of more baselines on \Bird \textit{w/o} oracle knowledge.}
    \label{tab:comparison}
\end{table}

\end{document}